\documentclass[letterpaper, 10 pt, conference]{ieeeconf}  

\IEEEoverridecommandlockouts              

\usepackage{times}

\usepackage{cite}
\usepackage{amsmath,amssymb,amsfonts}
\usepackage{algorithmic}
\usepackage{graphicx}
\usepackage{textcomp}
\usepackage{xcolor}
\usepackage{multirow}
\usepackage{subcaption}

\DeclareMathOperator*{\argmin}{arg\,min}
\usepackage[font=small,labelfont=bf]{caption}

\title{\LARGE \bf DMCA: Dense Multi-Agent Navigation using Attention and {\color{black}Selective} Inter-Agent Communication}

\author{Senthil Hariharan Arul$^{1}$, Amrit Singh Bedi$^{2}$ and Dinesh Manocha$^{3}$
\thanks{*This work was not supported by any organization}
\thanks{$^{1}$Author is with Dept. of Electrical and Computer Engineering, University of Maryland, College Park, MD, USA.
        {\tt\small sarul1@umd.edu}}%
\thanks{$^{2}$Author is with UMIACS, University of Maryland, College Park, MD, USA.
        {\tt\small amritb@umd.edu}}%
\thanks{$^{3}$Author is with Dept. of Computer Science, University of Maryland, College Park, MD, USA.
        {\tt\small dmanocha@umd.edu}}%
}

\begin{document}
\maketitle
\thispagestyle{empty}
\pagestyle{empty}

\begin{abstract}
In decentralized multi-robot navigation, {\color{black}ensuring safe and efficient movement with limited environmental awareness remains a challenge. While robots traditionally navigate based on local observations, this approach falters in complex environments. A possible solution is to enhance understanding of the world through inter-agent communication, but mere information broadcasting falls short in efficiency. In this work, we address this problem by simultaneously learning decentralized multi-robot collision avoidance and selective inter-agent communication. We use a multi-head self-attention mechanism that encodes observable information from neighboring robots into a concise and fixed-length observation vector, thereby handling varying numbers of neighbors.} Our method focuses on improving navigation performance through selective communication. We cast the communication selection as a link prediction problem, where the network determines the necessity of establishing a communication link with a specific neighbor based on the observable state information. The communicated information enhances the neighbor's observation and aids in selecting an appropriate navigation plan. By training the network end-to-end, we concurrently learn the optimal weights for the observation encoder, communication selection, and navigation components. We showcase the benefits of our approach by achieving safe and efficient navigation among multiple robots, even in dense and challenging environments. Comparative evaluations against various learning-based and model-based baselines demonstrate {\color{black}our} superior navigation performance, resulting in an impressive improvement of up to 24\% in success rate within complex evaluation scenarios.
\end{abstract}

\IEEEpeerreviewmaketitle

\section{Introduction}
Safe and efficient navigation of multiple robots is at the core of various robotic applications, including warehouse and factory automation, emergency response and rescue, natural resource monitoring, and outdoor industrial operations. Deploying multiple robots to operate simultaneously improves efficiency and throughput, making it beneficial for these applications. A critical issue in multi-robot navigation is designing collision-free trajectories for each robot. This problem has gathered significant attention over the past decade and is categorized into two primary classes of multi-robot algorithms, namely \emph{centralized} and \emph{decentralized}.

The centralized methods~\cite{tang2018complete,usc} view the robots as a single composite system, meaning the central system or server has global knowledge about all the robots. Centralized path generation has gained widespread application in warehouse automation due to the relative ease of guaranteeing efficient, collision-free paths. However, these methods have limited scalability owing to the central trajectory computation. In the worst case, their computation time can increase exponentially with the number of agents~\cite{solovey2016finding,goldenberg2014enhanced}. 

In decentralized navigation~\cite{berg2011reciprocal,van2008reciprocal,zhou2017fast}, robots make independent decisions using local sensing of the environment. {\color{black}Overall, the computation cost per agent is scalable, enabling large-scale deployment.} {\color{black}Recently, deep RL-based methods~\cite{cadrl,pedRich,marl1,marl2} have been developed for multi-agent navigation applications}. RL methods leverage their extensive offline training to learn and map the observation directly into actions. As with many deep learning techniques, they lack explainability and rigorous safety guarantees but have shown improved success rates and time-to-goal compared to prior rule-based or model-based methods on empirical evaluations. Despite their advantages, decentralized methods lack global knowledge about other agents and a central decision-maker. The distributed decision-making complicates reaching a consensus or cooperation between the agents and can result in less efficient paths, robot freezing behaviors, and even collisions in dense scenarios.

Human navigation is decentralized, where each person individually plans to avoid collision and navigate toward their goal. Moreover, humans can rely on  communication to improve their navigation decisions by compensating for their limited world knowledge. For example, while driving, humans indicate their intent using turn signal indicators to neighbors; when walking, they could request an individual to move and give way. This form of communication is explicit and intentional. Besides, humans can communicate their intent implicitly by looking in the direction of motion or changing their shoulder pose. Thus, humans navigating in a decentralized fashion can {\color{black}leverage} explicit or implicit communication to improve their navigation.

\subsection{Main Contributions}
{\color{black}Our approach is inspired from human navigation and 
leverages communication to improve navigation decisions in decentralized multi-robot settings.
We use a RL-based method to collectively learn multi-agent navigation, selective communication, and information aggregation. A key issue is to reduce the communication, as broadcasting information between robots can result in redundant information or can be costly in challenging scenarios with a large number of agents. Our approach uses novel RL techniques that can jointly learn multi-agent navigation and whom to communicate with. Our main contributions include:}

\begin{enumerate}
    \item We present {\color{black}DMCA}, a novel approach {\color{black}that leverages selective inter-agent communication to enhance decentralized} multi-robot navigation in dense scenarios. We use deep reinforcement learning to simultaneously learn collision-free navigation, ``whom to communicate {\color{black}information", and information aggregation,} which improves the overall navigation performance.

    \item To deal with a variable number of neighbors for each robot, we utilize a multi-head self-attention mechanism to encode the observation vectors from all neighbors to create a fixed-length observation vector. 

    \item {\color{black}We formulate selective communication as a link prediction problem, where our network predicts the probability of a communication link with the robot based on the neighbors' observed state information. Moreover, the selectively communicated information gets aggregated using an LSTM to a fixed-length encoding which augments the robot's locally observed information. Our network is observed to communicate considerably lower than broadcast (2-10x lower in certain cases) while improving the navigation performance over compared baseline methods.}
\end{enumerate}

We evaluate our method on multiple simulated benchmarks (with upto 30 agents) to compare its navigation performance against prior learning-based methods (CADRL~\cite{cadrl}, GA3C-CADRL~\cite{pedRich}, and~\cite{long}) and model-based methods (ORCA~\cite{berg2011reciprocal}, BVC~\cite{zhou2017fast}, and BUAVC~\cite{zhu2022decentralized}). We consider metrics such as path length, time-to-goal, collision rate, and the overall success rate in reaching the goal. We observe up to $24\%$ improvement in success rate over the chosen baseline methods in different scenarios. {\color{black}Moreover, we present a discussion on the performance of DMCA in terms of the communication links predicted, which shows significantly reduced communication than broadcasting.} On average, our method takes $\sim6$ ms per agent to compute the navigation action with tens of agents in our benchmarks. 
\section{Related Works}

{\subsection{Model-Based {\color{black}Decentralized} Collision Avoidance}}
{\color{black}Velocity Obstacle}~\cite{fiorini1998motion} computes collision-free velocities for the agents based on their observations of neighbors' positions and velocities. RVO~\cite{van2008reciprocal} improves on VO by incorporating reciprocity in terms of velocity computation. ORCA~\cite{berg2011reciprocal} linearizes the RVO constraints to improve computational efficiency and is generalized to different agent dynamics in~\cite{bareiss2015generalized}. {\color{black}BVC~\cite{zhou2017fast} constructs a Voronoi-based free space for each agent to perform collision-free navigation.} V-RVO~\cite{vrvo} presents a hybrid between RVO and BVC for improved collision avoidance performance and passive safety guarantees. Traditionally, decentralized model-based methods have shown impressive performance in terms of safety, but they often become conservative and drive the robots to a deadlock affecting the success rate.

{\subsection{Learning-Based Collision Avoidance}}
The idea of utilizing deep RL to achieve better navigation performance has been investigated in the literature. CADRL~\cite{cadrl} proposed an RL method for multi-agent navigation, which showed improved time-to-goal performance compared to ORCA. Semnani et al.~\cite{hybridFMP} presented a hybrid framework that switches between RL and force-based planning based on the scene complexity, resulting in an improved success rate and time-to-goal. Everett et al.~\cite{pedRich} further improved CADRL to account for an arbitrary number of neighboring agents by using an LSTM to encode a varying-size observation vector into a fixed-length vector. In~\cite{chen2019crowd}, a self-attention mechanism is used to aggregate pairwise interactions. In contrast, our method uses multi-head self-attention~\cite{attention} to encode the observation to a fixed-length vector. Self-attention is independent of sequence order, unlike LSTM. Long et al.~\cite{longDRL} learn to map raw sensor data to action. The method shows a better success rate, path optimality, agent's average speed, and time-to-goal than NH-ORCA~\cite{nhorca}. Fan et al.~\cite{FanHybridRL} further improved this performance by adopting a hybrid framework. Mapping raw sensor data directly to action can significantly suffer from a sim-to-real performance gap. Xu and Karamouzas~\cite{xu2021humaninspired} used expert human trajectories and knowledge distillation to shape the reward and generate human-like trajectories. Further, the authors showed improved energy efficiency and success rate compared to ORCA. Tan et al.~\cite{Qingyang} propose using both global and local information about the environments to navigate and show improved performance in dense scenarios. GLAS~\cite{GLAS} presented a distributed, provably safe policy generation for multi-agent planning that uses globally planned trajectories, constructs a local observation action training set, and is used to learn a decentralized policy using deep imitation learning. The above two works show the benefits of global information for improved navigation performance. Our methods, inspired by communication in human navigation and the advantages of enhancing world knowledge, use selective inter-robot communication to improve world knowledge for navigation. {\color{black}DMCA communicates considerably lower with their neighbors compared to a broadcast type communication.}

DRL-VO~\cite{drlvo} uses a velocity obstacle (VO) based cost in the rewards to improve the success rate. Han et al.~\cite{han2022reinforcement} presented a DRL method with an RVO-shaped reward for better reciprocal behavior. Li et al.~\cite{li2020reciprocal} proposed a hybrid method based on RL and ORCA. The RL network computes the desired action for each neighboring agent, weighted to compute a suitable preferred velocity used in the ORCA formulation. A safety module based on the control barrier function (CBF) trains the network end-to-end to ensure collision-free navigation. Cai et al~\cite{cai2021safe} propose CBF-based shielding for safety-critical MARL tasks. 

{\subsection{Learning Communication}}
Recently, learning communication has received significant attention. IC3Net~\cite{singh2018learning} employs a gating mechanism to learn when to communicate. SchedNet~\cite{kim2019learning} allows selected agents to broadcast information based on observations from all agents. {\color{black}In Variance-based Control (VBC)~\cite{VBC}, each agent decides to broadcast a message or reply based on the variance produced by the action.} DGN~\cite{DGN} uses multi-head attention to gather information from the neighborhood. TarMAC~\cite{Tarmac} uses attention to collect relevant messages in a broadcast setting.  

In I2C~\cite{NEURIPS2020_fb2fcd53}, authors address the problem of whom to communicate with, but I2C relies assumptions such as joint observation space between agents which can be unsuitable for a decentralized navigation. ToM2C~\cite{tom2c} uses the theory of mind model to infer the agent's intention and to decide when and with whom to communicate.

A few works have explored learning communication in the multi-agent navigation domain. Serra-Gomez et al.~\cite{wwtc} learn with whom to communicate and request the planned trajectories from the chosen neighbor to be used in the MPC planner. Ma et al.~\cite{ma_broadcast} present a DRL method for multi-agent pathfinding with broadcast communication. In~\cite{ma2021learning}, they reduce the communication overhead by combining the idea from \cite{NEURIPS2020_fb2fcd53} for multi-agent navigation in a grid world domain. In contrast, our method considers complex, non-grid world environments. Our network includes a communication module that communicates with select neighbors, improving the overall navigation performance. 
\section{Problem Formulation and Overview}
In this section, we list our assumptions, summarizes our problem statement, and provides an overview of our approach. In Table~\ref{tab:symbols}, we list the variables and notations frequently used in our paper.

\subsection{Assumptions}
In this paper, we assume disk-shaped robots with unicycle dynamics. We consider a request-reply type of communication between the ego-agent and its neighbors, i.e., an agent requests information from its neighbor, and the neighbor replies with goal information. Also, we utilize ego-agent to refer to the agent being considered. We assume the request-reply is fast and executed within a single planning cycle.
\begin{table}[t]
    \centering
    \scalebox{0.9}{

    \begin{tabular}{c|c}
        \hline
        Symbols & Description\\
        \hline\hline
        $\mathbf{p}_i$ & Position of agent i ($p_x,p_y$) \\
        $\mathbf{v}_i$ & Velocity of agent i ($v_x,v_y$) \\
        $\psi_i$ & Heading of agent i \\
        $r_i$ & Radius of agent i\\
        $v_{pref_i}$ & Preferred Velocity of agent i\\
        $\mathbf{g}_i$ & Goal of agent i ($g_x,g_y$) \\
        $\mathcal{N}_i$ & Set of neighbors of agent i \\
        $\mathcal{C}_i$ & Set of selected neighbors for communication\\
        $d_a$ & Inter-agent distance\\
        {\color{black}$\lambda_{comm.}$} & {\color{black}Weight for communication reward}\\
        \hline
    \end{tabular}}
    \caption{Symbols}
    \label{tab:symbols}
\end{table}
\subsection{Problem Statement}
We consider the problem of cooperative multi-robot navigation, where robots can communicate with each other. We propose a deep RL method to navigate individual robots toward their respective goals while avoiding collisions. Each robot communicates with selected neighbors to obtain information to improve navigation decisions. Mathematically, let us consider an environment $\mathcal{W} \subset \mathbb{R}^2$ with $N$ communicating robots $\{ \mathcal{A}_1, \mathcal{A}_2, \cdot \cdot \cdot, \mathcal{A}_n \}$. Each robot is a disk-shaped robot with 2D position $\mathbf{p_i}\in \mathcal{W}$, radius $r_i$, and goal $\mathbf{g}_i\in \mathcal{W}$. Let $t$ denote the time step, then we can express the per-step safe-navigation problem for robot $i$ as follows: 
\begin{align}
\argmin_{\pi} \quad & ||\mathbf{p}_i(t) - \mathbf{g}_i||\\
\quad & \Vert \mathbf{p}_{i}(t) - \mathbf{p}_{j}(t) \Vert_2 \ge r_i + r_j\\
  & \forall i,j \in \{1,2, ..., N\}, i\neq j \quad \forall t.
\end{align}
Here, $\mathbf{p}_i(t)$ represents the path of the robot $i$ as a function of time $t$. $\pi$ is the policy navigating the robot. Following the definition from CADRL~\cite{cadrl} , the agent's state $\mathbf{s}_i = [\mathbf{s}_i^o, \mathbf{s}_i^h]$ includes an observable component and a hidden component. The observable component $\mathbf{s}_i^o$ includes the agent's position, velocity, and radius. The hidden component $\mathbf{s}_i^h$ includes the agent's goal, preferred speed, and current orientation. Hence, we represent them as:
\begin{align*}
    \mathbf{s}_i^o = [{p}_x, {p}_y, {v}_x, {v}_y, r], \quad
    \mathbf{s}_i^h = [{g}_x, {g}_y, v_{pref}, \psi].
\end{align*}
The agent's control input includes its speed and heading angle and is given by $\mathbf{u}_i = [v_i, \psi_i]$.

\subsection{RL for Multi-Robot Navigation}
A relevant metric to evaluate the navigation performance is the success rate, which is the fraction of the total number of robots reaching the goal configuration collision-free and deadlock-free. As mentioned earlier, the limited world knowledge and distributed decision-making complicate the navigation problem. 
Traditionally decentralized model-based methods have shown impressive performance in multi-robot navigation scenarios. They can be designed to ensure safety, but they often can become conservative and drive the robots to a deadlock affecting the success rate.
In this paper, we use RL to improve the success rate (and navigation performance) by leveraging extensive offline training. Using RL for navigation opens a few challenges. Firstly, the network is required to handle a varying number of neighbors. Secondly, the goal of communication is to enhance navigation behavior and should be learned simultaneously.

\subsection{Multi-head Attention}
Initially proposed in~\cite{attention}, self-attention has shown immense potential in natural language processing and computer vision. Self-attention mechanisms compare elements of an input sequence with each other to compute a representation of the sequence. For each element in the input sequence, the attention mechanism calculates Query (Q), Key (K), and Value (V) vectors by multiplying them with trained matrices. Using the Q, K, and V vectors, a self-attention score is computed for each element in the sequence. The score determines the focus awarded to other sequence elements for encoding a particular element, we can write
$$Attention(Q,K,V) = \bigg ( \frac{QK^T}{\sqrt d_k}\bigg )V, $$
where, $\sqrt d_k$ is the dimension of K.
In multi-head attention~\cite{attention}, multiple heads are created by linearly projecting Q, K, and V. Each self-attention head focuses on different subspaces, and attention is performed in parallel. 
The projected queries, keys, and values are fed into attention pooling in parallel. Next, attention pooling outputs are concatenated and transformed with another learned linear projection to produce the final output. So, we can write
$$\text{Multi-head(Q,K,V)} = \text{Concat}(head_1, head_2, ... head_k)W^O,$$
$$    \text{head}_i = \text{Attention}(Q W_i^Q,K W_i^K,V W_i^V),$$
where $W_i^Q, W_i^K, W_i^V, W^O$ are parameter matrices for projection.
Prior works, like~\cite{pedRich}, use LSTM to encode the observation, but LSTMs are known to focus more on the recent input. Using self-attention can avoid the encoding being order dependent. In our case, the input sequences are the neighbors' observable states. As self-attention compares the elements of a sequence with each other, we include an element pertaining to the ego agent in the sequence. 

\subsection{Inter-agent Communications}

Communication in a multi-agent setting involves transferring information such as state, intent, or observations between the agents. Primarily, communication is categorized into broadcast or selective. In broadcast, an agent publishes its information to be used by all agents. While in selective communication, the agents make intelligent decisions about when and with whom to communicate, improving communication efficiency. In our formulation, the agent communicates their intents by transferring their hidden state with their neighbors.

\subsection{Reinforcement Learning Based Solution} \label{RLinfo}
As with prior methods~\cite{cadrl,pedRich}, we consider a local coordinate frame with the state composed of information about the ego-agent and its neighbors. The information about the ego-agent includes its distance to the goal, preferred speed, orientation, and radius as
\begin{equation}
    \mathbf{s}^{ego} = [||\mathbf{g}_i - \mathbf{p}_i||,\;\; v_{pref_i},\;\; \psi_i, \;\; r_i].
\end{equation}
The world information includes the agent's neighbors, including their positions, velocities w.r.t the ego frame, radii, inter-agent distances, and combined radii denoted as 
\begin{align}
\mathbf{s}^{obs_j} = [\Tilde{p_{x_j}},\; \Tilde{p_{y_j}},\; \Tilde{v_{x_j}},\; \Tilde{v_{y_j}},\; r_j,\; d_a,\; r_i+r_j],\\
\mathbf{s}^{obs} = [\mathbf{s}^{obs_1}, \mathbf{s}^{obs_2}, \cdot\cdot\cdot, \mathbf{s}^{obs_n}]
\quad {1,2, ..., n} \in \mathcal{N}_i.
\end{align}
In addition, our method allows the agent to request one or more of its neighbors for their hidden state ($s_j^h$) using communication to augment the network's input. The communicated information is used to construct a communicated state from each neighbor and includes distance to a neighbor's goal from the ego-agent, the difference in preferred speeds between the ego-agent and its neighbor, and relative orientation/heading. More details are presented in Section~\ref{sec:comm}.

Based on the prior approach of~\cite{pedRich}, our multi-agent RL problem formulation is trained with GA3C and uses a similar navigation reward. We positively reward the agents for reaching the goal and negatively reward them for a collision and close proximity to other agents. $d_{min}$ is the distance to the closest neighbor. {\color{black}In addition, we have a negative reward for every communication link ($\lambda_{comm.}$). The reward structure is shown below.
\begin{equation}
R =
    \begin{cases}
      1.0, & p_i = g_i\\
      -0.25, & \text{collision}\\
      -0.1 + d_{min}/2 &\text{if } d_{min} < 0.2\\
      -\lambda_{comm.}\cdot n_l & \text{$n_l$: no of comm. links}\\
      0.0, & \text{otherwise}
    \end{cases}
\end{equation}
}

\section{{\color{black}Proposed Approach: DMCA}}
\begin{figure*}[h]
\centering
\includegraphics[width=0.7\linewidth, trim={10cm, 0cm, 0.5cm, 0cm}, clip]{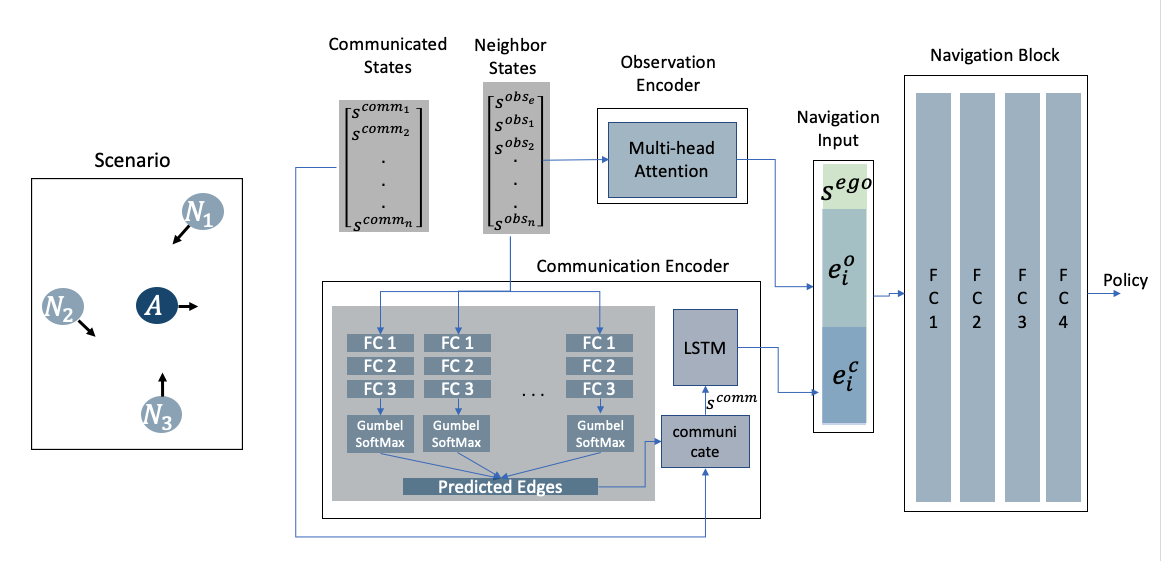}
\caption{We illustrate the high-level network architecture used for multi-agent navigation in DMCA. Primarily, the network consists of three modules: the observation encoder, the communication selection, and the navigation block.}\label{fig:framework}
\end{figure*}
In this section, we illustrate our proposed network architecture (Figure~\ref{fig:framework}) and describes its  modules. 
Our network consists of 3 components: the observation encoder, the communication module, and the navigation module. The observation encoder performs the task of encoding neighbors' information for use in the policy. The communication module links with neighbors to obtain their hidden states, resulting in more information about the world for the ego-agent. The navigation module maps the encoded vectors to a distribution over the action space.

\subsection{Observation Encoding}
The observation encoder takes the observation vector of all neighboring robots ($\mathbf{s}^{obs_j}$) as input to create a fixed-length vector for input to the navigation module. Since the number of neighbors around an agent varies at any point in time, our method needs to account for varying numbers of agents. Thus, our observation encoder uses a multi-head self-attention to encode the neighbors $\mathbf{s}^{obs_j}$ into a fixed-length observation vector. 

The input sequence consists of $\mathbf{s}^{obs_j}$ for each neighbor, which is represented relative to the ego agent. In addition, we compute $\mathbf{s}^{obs_{ego}}$, which is the observed state of the ego agent relative to the ego frame. The ego-agent's observed state is needed for our self-attention mechanism. Based on the ego observed state, we compute the attention paid to neighboring observed states, which is combined to produce the encoded vector. The input sequence consists of the ego agent's observable state, followed by its neighbors' observable state. Finally, the representation computed for $\mathbf{s}^{obs_{ego}}$ is used as the encoded representation.
\begin{equation}
    \mathbf{e}^{o} = \texttt{encode}([\mathbf{s}^{obs_{ego}}, \mathbf{s}^{obs}])
\end{equation}

Our encoder uses 20 heads for our observation encoding module, with Key, Query encoding using a dense layer with 128 nodes; the Value has 256 nodes. 

\subsection{Communication Selection}\label{sec:comm}
This block performs communication selection using the robot's local observation. We formulate the communication selection as a link prediction problem between the ego robot and its neighbors. The set of neighbors for an agent $i$ is given by
$
    \mathcal{N}^i = \{j \quad | \quad j\neq i, \; ||\mathbf{p}_j - \mathbf{p}_i|| < r_{neighbor}\},
$
where $r_{neighbor}$ represents a radius threshold.
The module takes the neighbor's observable states as input and individually passes them through a series of three fully connected layers. Since $s^{obs_j}$ is in relative frame w.r.t to the ego agent, we pass the vector through a sequential layer to predict the possibility of a communication link.

In this regard, the output of the sequential layer has 2 nodes, one indicating the probability of a link ($p_{link}$) and the other node indicating $1-p_{link}$. The first 2 layers have 64 nodes and relu activation, with the last layer having 2 output nodes and softmax activation. The Gumbel-Softmax~\cite{gumbel1,gumbel2} layer is used to sample a discrete distribution based on the probability of a link. 
\begin{equation}
    \forall j \in \mathcal{N}^i \quad [p_{link_j}, 1-p_{link_j}] = \texttt{Comm.}\; \texttt{Sel}(s^{obs_j}).
\end{equation}
Thus, for each neighbor, the communication module predicts a communication link. If a link is predicted, the agent sends a communication request to the selected neighbors. The neighbors respond with the hidden states of the agents $\mathbf{s}^{h} = [{g}_x, {g}_y, v_{pref}, \psi]$. 

The received hidden states are combined with the observed states from the neighbors to create a communication state. The communication state is given by,
\begin{equation}
\mathbf{s}^{comm_j} = [||\mathbf{g}_j - \mathbf{p}_i||, \;\; v_{pref_j} - v_{pref_i},\;\; \psi_j - \psi_i] \quad j \in \mathcal{N}_i
\end{equation}
\begin{equation}
\mathbf{s}^{comm} = \bigg[ [\mathbf{s}^{obs_1}, \mathbf{s}^{comm_1} ],\cdot\cdot\cdot,
[ \mathbf{s}^{obs_n}, \mathbf{s}^{comm_n} ] \bigg],\; {1 \text{ to } n} \in \mathcal{C}_i.
\end{equation}
We pass the communication vector through an LSTM to create an encoded vector $\mathbf{e}^c = \texttt{LSTM}(\mathbf{s}^{comm_j}).$
\subsection{Navigation Input}
The navigation input is the input to the navigation module and consists of three important vectors: the host agent state, the encoded observation vector, and the encoded communication vector. That is, 
$
\mathbf{s}^{input} = [\mathbf{s}^{ego}, \mathbf{e}^o, \mathbf{e}^c].
$
\subsection{Navigation}
Our navigation module consists of a sequential layer with 4 fully connected layers. The first layer has 1024 nodes, the next two layers have 512 nodes, and the last layer has 256 nodes.
As in Everett et al.~\cite{pedRich}, the output from the final layers includes the scalar value function and the distribution over the action space.
\section{Evaluation}~\label{sec:eval}
\begin{figure*}[t]
\centering
  \begin{subfigure}[b]{.19\linewidth}    
        \includegraphics[width=\textwidth, trim={1cm, 1cm, 0cm, 1cm}, clip]{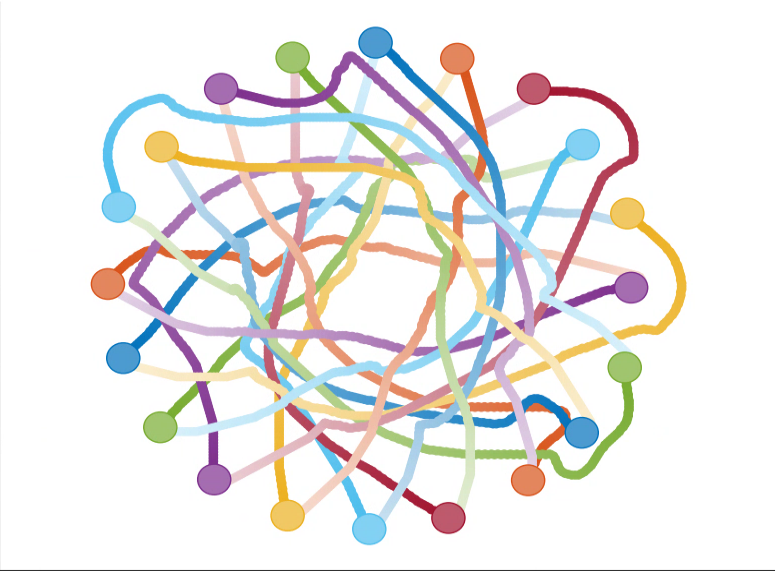}
        \caption{DMCA}
        \label{img1}
 \end{subfigure}
  \begin{subfigure}[b]{.19\linewidth}    
        \includegraphics[width=\textwidth, trim={1cm, 1cm, 0cm, 1cm}, clip]{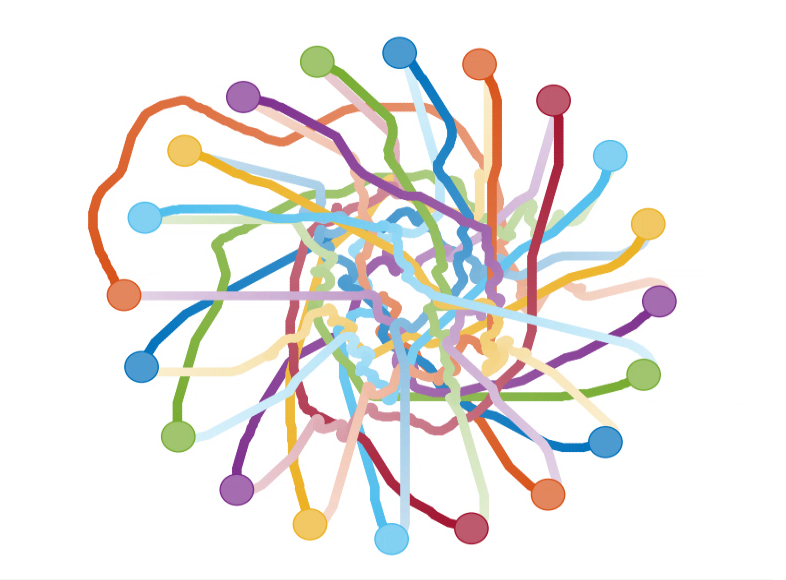}
        \caption{DMCA-LC}
        \label{img1}
 \end{subfigure}
 \begin{subfigure}[b]{.19\linewidth}  
        \includegraphics[width=\textwidth, trim={1cm, 1cm, 0cm, 1cm}, clip]{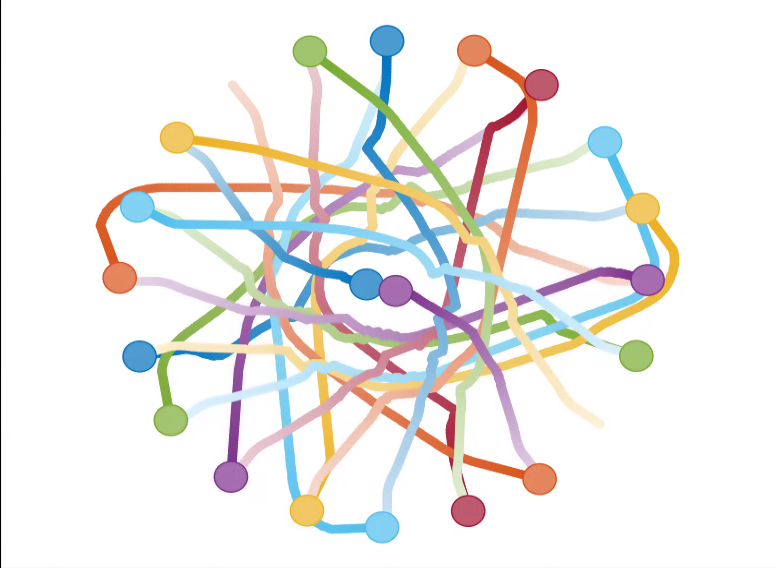}
        \caption{CADRL}
        \label{img3}
 \end{subfigure} 
  \begin{subfigure}[b]{.19\linewidth}  
        \includegraphics[width=\textwidth, trim={1cm, 1cm, 0cm, 0.8cm}, clip]{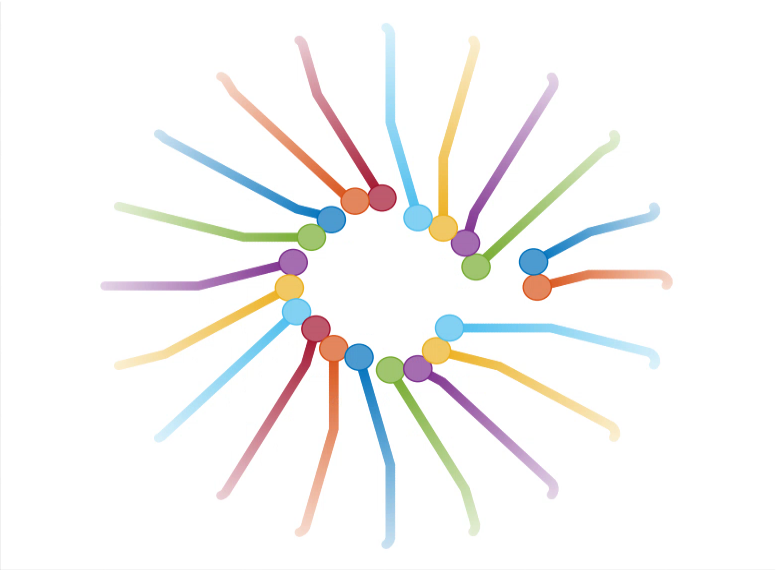}
        \caption{GA3C-CADRL}
        \label{img2}
 \end{subfigure}
   \begin{subfigure}[b]{.20\linewidth}  
        \includegraphics[width=\textwidth, trim={6cm, 6cm, 6cm, 4cm}, clip]{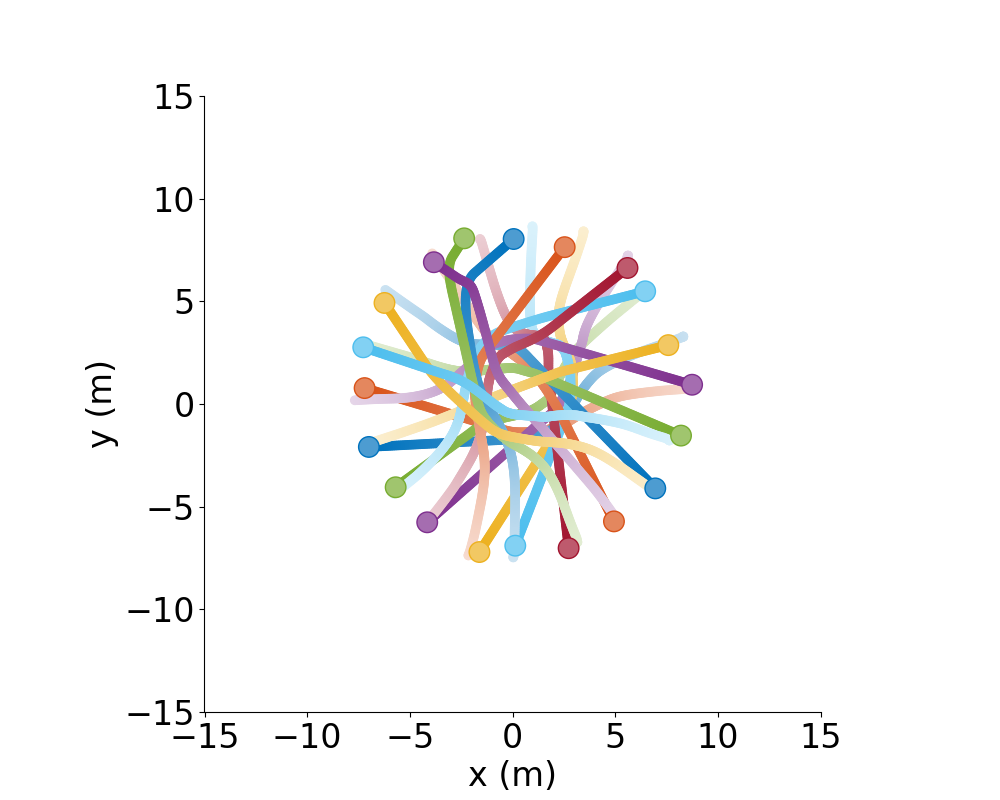}
        \caption{ORCA}
        \label{img2}
 \end{subfigure}
    \caption{We compare the trajectories generated by our proposed method with CADRL and GA3C-CADRL for a circular scenario with 20 agents. We observe that DMCA generates smooth and collision-free trajectories to the goal, while CADRL results in some collisions. In GA3C-CADRL, some agents were deadlocked while others were in a collision, and no agent reached the goal. The network requires ~6 ms per agent to compute an action.}\label{fig:traj}
\end{figure*}

\begin{figure*}
\centering
   \begin{subfigure}[b]{.19\linewidth}    
        \includegraphics[width=0.95\textwidth, trim={1cm, 3cm, 0cm, 2cm}, clip]{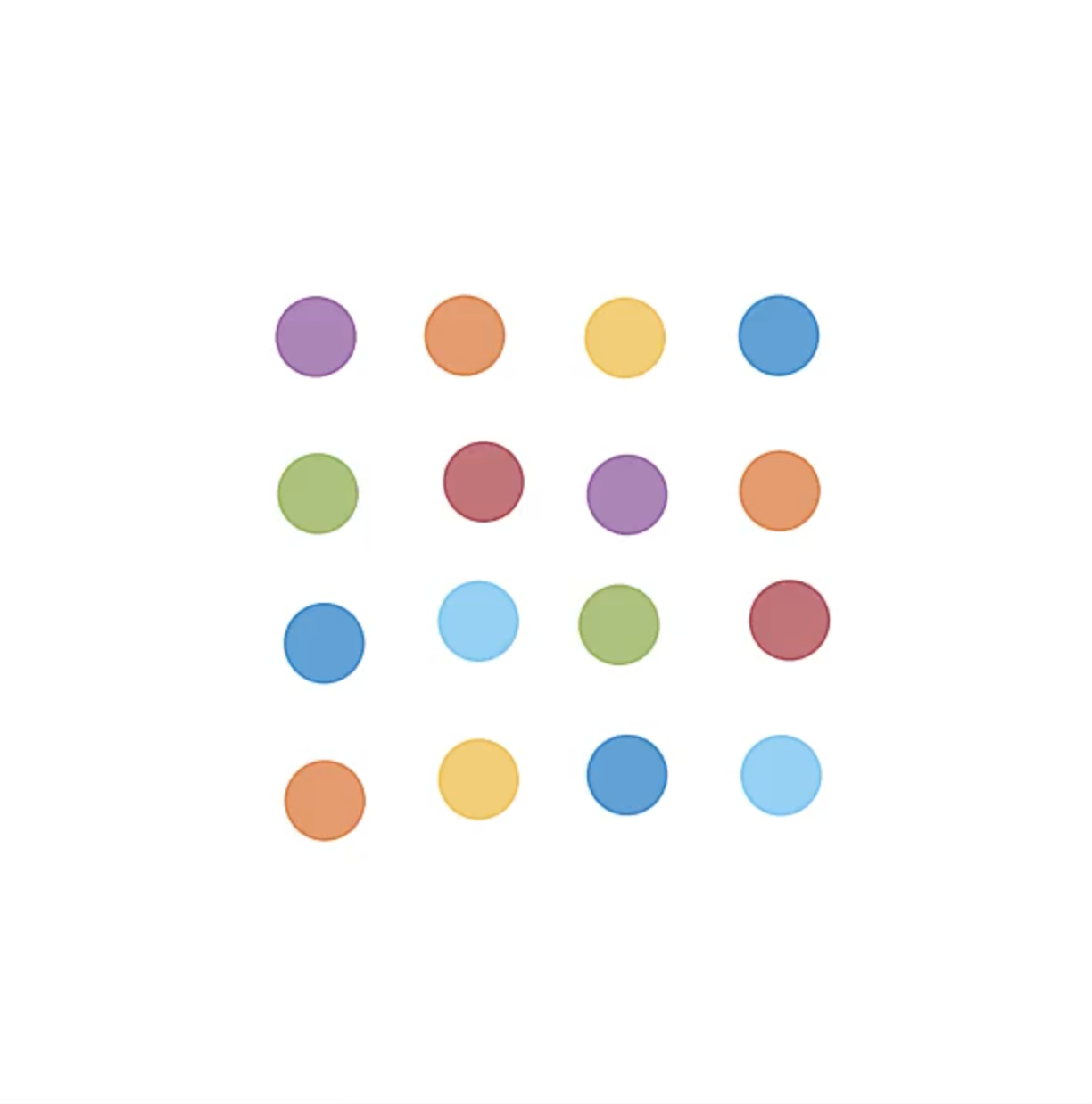}
        \caption{Initial Configuration}
        \label{init}
 \end{subfigure}
   \begin{subfigure}[b]{.19\linewidth}    
        \includegraphics[width=0.95\textwidth, trim={1cm, 3.0cm, 0cm, 2cm}, clip]{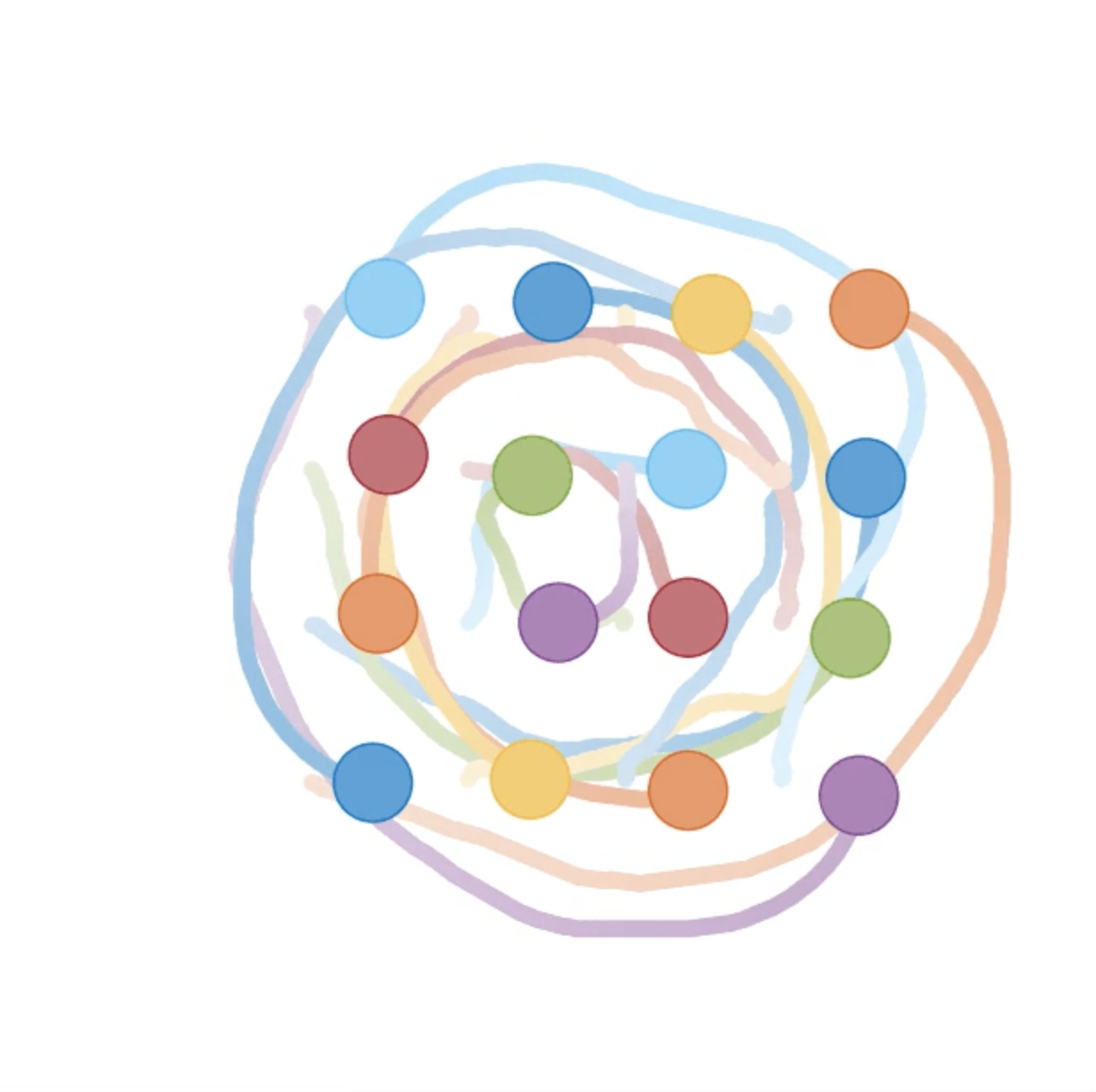}
        \caption{DMCA}
        \label{NavSIC}
 \end{subfigure}
   \begin{subfigure}[b]{.19\linewidth}    
        \includegraphics[width=0.95\textwidth, trim={1cm, 3cm, 0cm, 2cm}, clip]{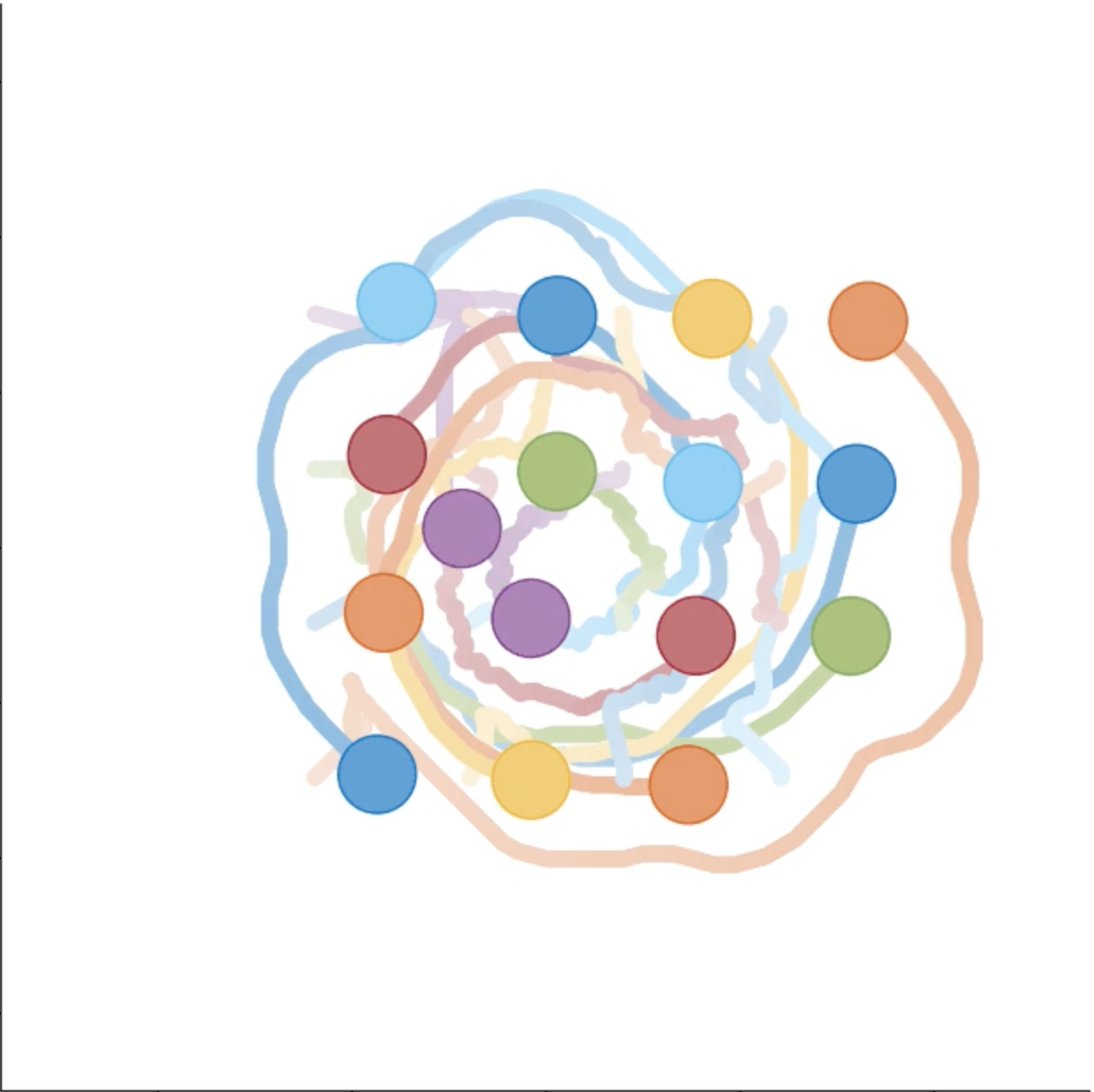}
        \caption{DMCA-LC}
        \label{NavSIClc}
 \end{subfigure}
   \begin{subfigure}[b]{.19\linewidth}    
        \includegraphics[width=0.95\textwidth, trim={1cm, 3cm, 0cm, 2cm}, clip]{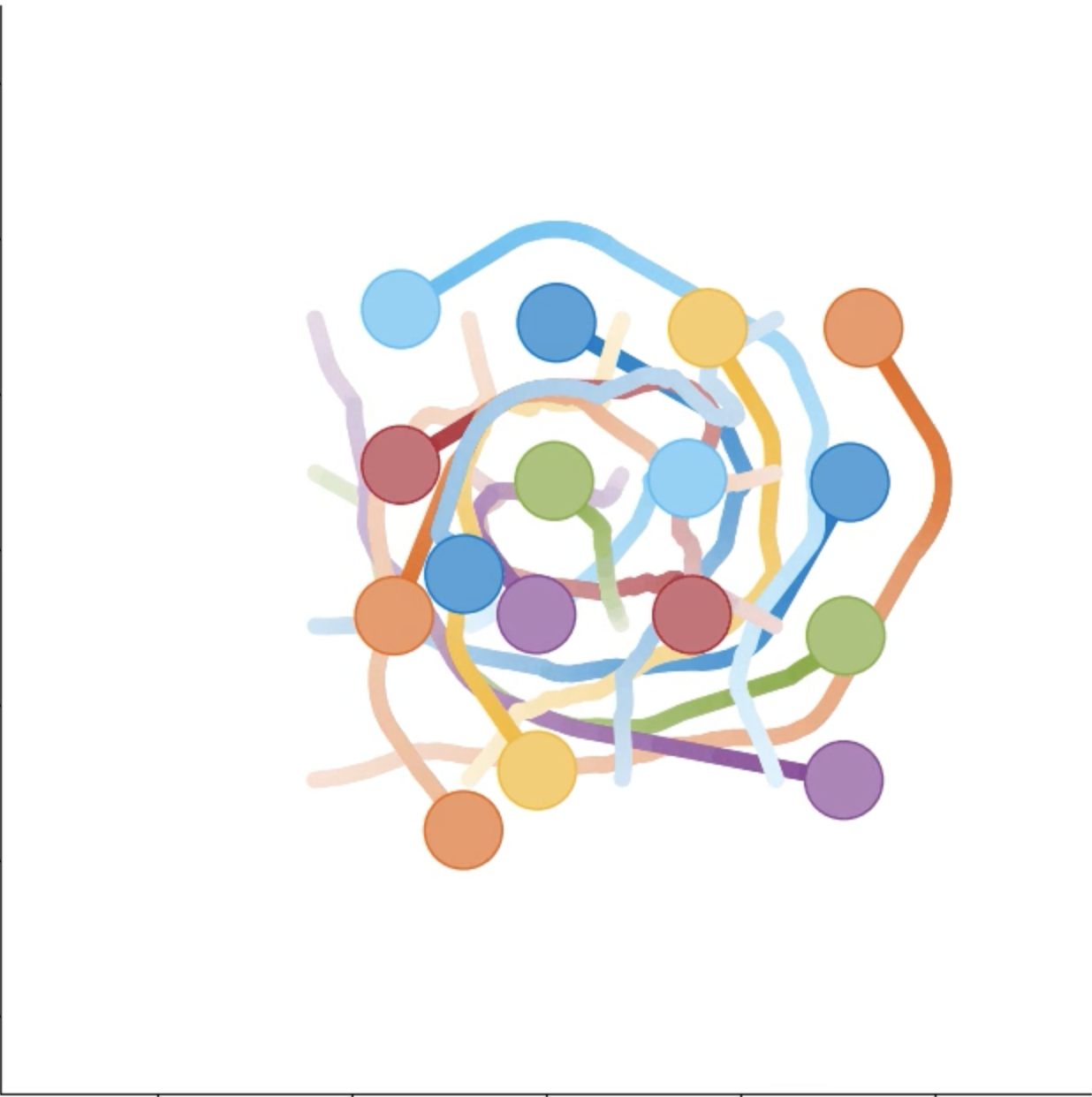}
        \caption{CADRL}
        \label{cadrl}
 \end{subfigure}
   \begin{subfigure}[b]{.19\linewidth}    
        \includegraphics[width=0.95\textwidth, trim={1cm, 3cm, 0cm, 2cm}, clip]{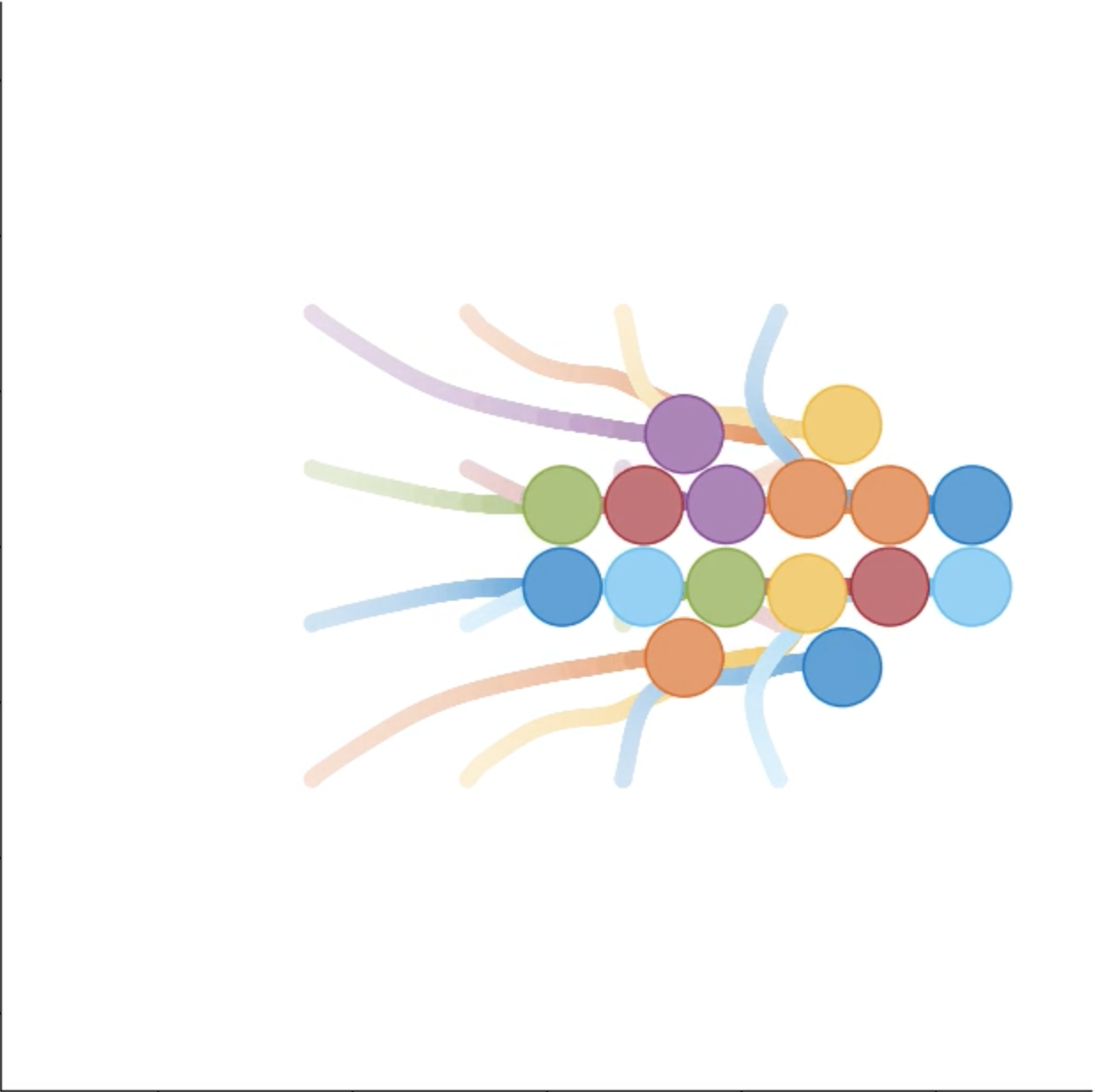}
        \caption{ORCA}
        \label{RVO}
 \end{subfigure}
    \caption{We consider a complex scenario where the robots are arranged in an $n \times n$ grid formation. The final formation is created by moving the robot at position $(\texttt{row},\texttt{column})$ to $(\texttt{n-row+1}, \texttt{n-column+1})$. Figure~\ref{init} shows the initial configuration of the agents, Figure~\ref{NavSIC}-~\ref{RVO} shows the final configuration for DMCA, DMCA-LC, CADRL, and ORCA. We observe that DMCA generates collision-free trajectories in this scenario while DMCA-LC results in one robot being deadlocked. CADRL results in some robots colliding, while results in agents colliding or deadlocked. Thus DMCA and DMCA-LC perform the best in this scenario. }\label{fig:traj_form}
\end{figure*}
\begin{figure*}[t]
\centering
   \begin{subfigure}[b]{.24\linewidth}    
        \includegraphics[height=2.2cm, width=0.9\textwidth, trim={0.1cm, 0.1cm, 0cm, 1cm}, clip]{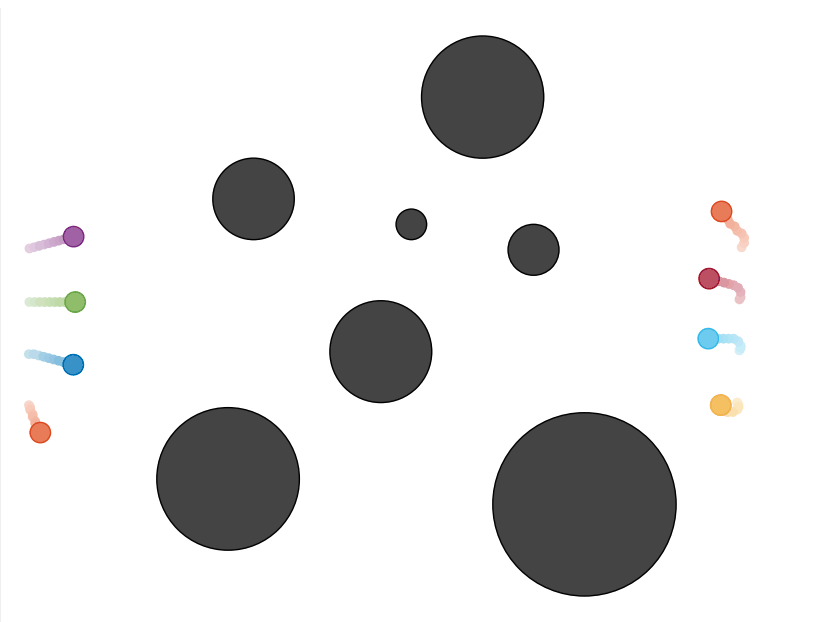}
        \label{img1}
 \end{subfigure}
   \begin{subfigure}[b]{.24\linewidth}    \includegraphics[height=2.2cm,width=0.9\textwidth, trim={0.1cm, 0.2cm, 0cm, 1cm}, clip]{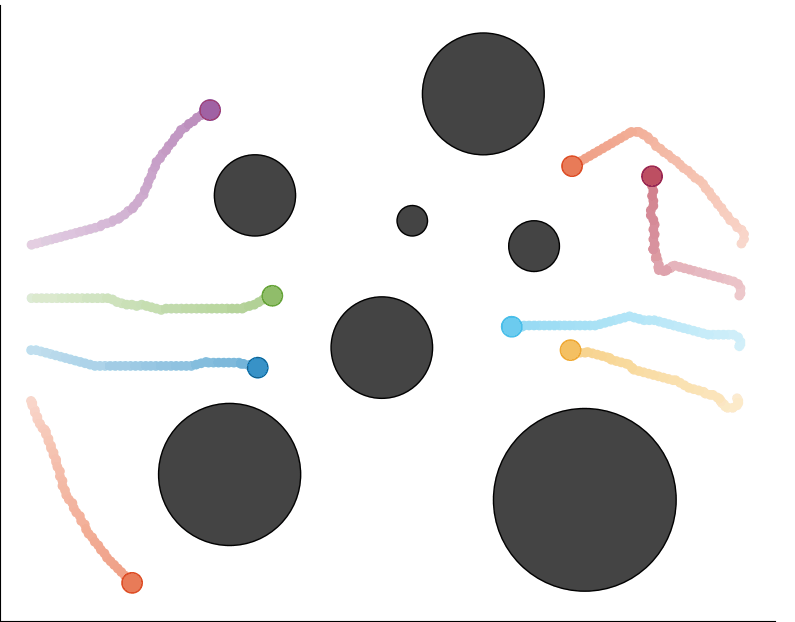}
        \label{img1}
 \end{subfigure}
   \begin{subfigure}[b]{.24\linewidth}    \includegraphics[height=2.2cm,width=0.9\textwidth, trim={1cm, 1cm, 0cm, 1cm}, clip]{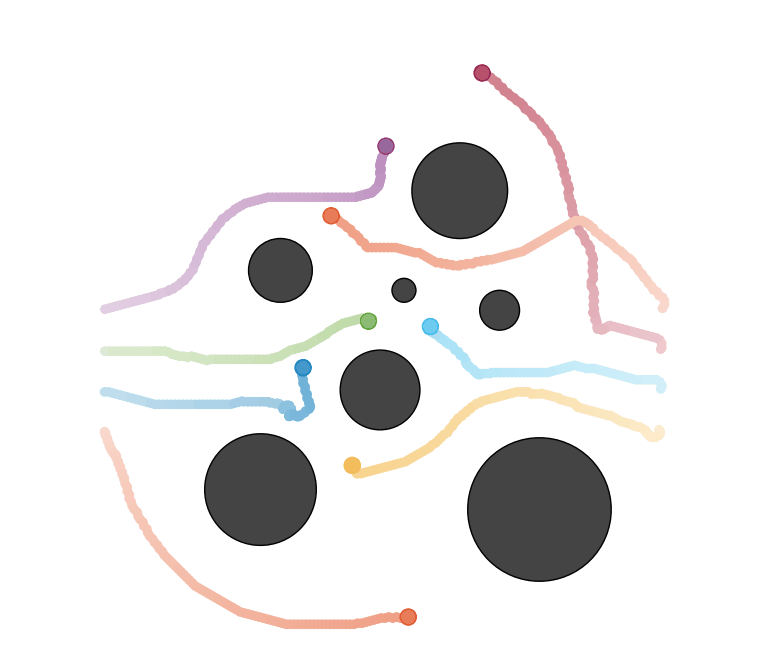}
        \label{img1}
 \end{subfigure}
   \begin{subfigure}[b]{.24\linewidth}    \includegraphics[height=2.2cm,width=0.9\textwidth, trim={1cm, 1cm, 0cm, 1cm}, clip]{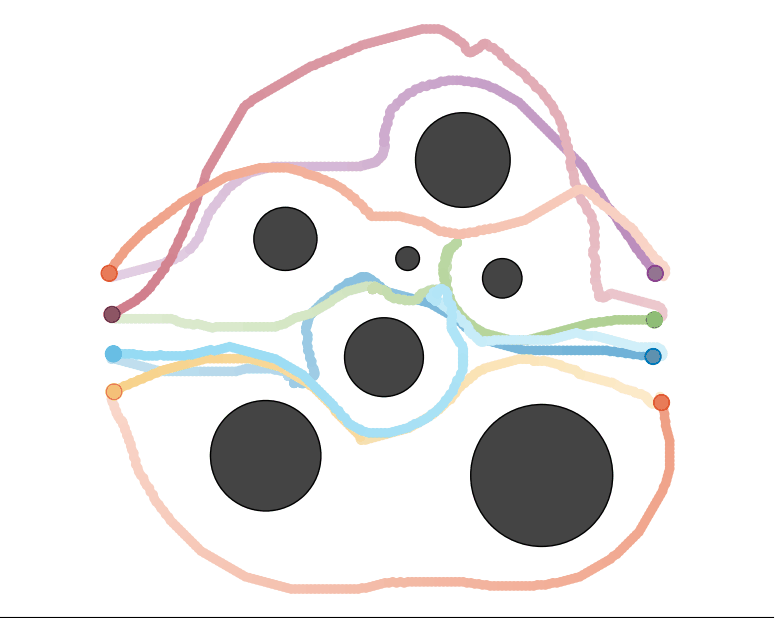}
        \label{img1}
 \end{subfigure}
    \caption{We consider a swap scenario with disk-shaped static obstacles (black) and eight agents. We observe that the eight agents using DMCA remain safe and successfully reach their goal in this scenario.}\label{fig:traj_static_2}
\end{figure*}

\begin{table*}
    \centering
    \scalebox{0.99}{
    \begin{tabular}{|c|c|c|c|c|c||c|c|c|c|c||c|c|c|c|c|c|}
        \hline
        \multirow{2}{*}{Robots} & \multicolumn{5}{c||}{Collision Rate} & \multicolumn{5}{c||}{Time to Goal} & \multicolumn{5}{c|}{Success Rate (SR)} & \multirow{2}{*}{Avg. SR} \\
        \cline{2-16}
         & 4 & 6 & 10 & 20 & 30 & 4 & 6 & 10 & 20 & 30 & 4 & 6 & 10 & 20 & 30 & \\
        \hline
        \multicolumn{17}{|c|}{Scenario 1: Circle Scenario (agent radius 0.2m)}\\
        \hline
        NavSIC & 0.00 & 0.00 & 0.00 & 0.00 & 0.05 & 113.2 & 114.6 & 114.2 & 201.6 & 242.3 & 1.00 & 1.00 & 1.00 & 1.00 & 0.95 & 0.99\\
        \hline
        NavSIC-LC & 0.00 & 0.00 & 0.00 & 0.43 & 0.45 & 144.3 & 125.2 & 124.2 & NR & NR & 1.00 & 1.00 & 1.00 & 0.57 & 0.55 & 0.82\\
        \hline
        ORCA & 0.00 & 0.00 & 0.00 & 0.02 & 0.50 & 109.2 & 116.2 & 118.4 & 214.8 & NR & 1.00 & 1.00 & 1.00 & 0.98 & 0.36 & 0.87\\
        \hline
        CADRL & 0.00 & 0.00 & 0.00 & 0.00 & 0.00 & 108.0 & 111.5  & 126.3 & 201.2 & 212.8 & 1.00 & 1.00 & 1.00 & 1.00 & 0.99 & 0.99\\
        \hline
        GA3C & 0.10 & 0.30 & 0.62 & 0.90 & 0.97 & 108.3 & 109.8 & NR & NR & NR & 0.90 & 0.70 & 0.37 & 0.10 & 0.03 & 0.42\\
        \hline
        BVC & 0.00 & 0.00 & 0.00 & 0.00 & 0.00 & 222.5 & 227.2 & 345.4 & 265.2 & NC & 1.00 & 1.00 & 1.00 & 1.00 & NC & -\\
        \hline
        BUAVC & 0.00 & 0.00 & 0.00 & 0.00 & 0.00 & 181.2 & 226.5 & 408.3 & NC & NC & 1.00 & 1.00 & 1.00 & NC & NC & - \\
        \hline
        Long et al.\cite{long} & 0.00 & 0.00 & 0.00 & 0.04 & 0.12 & 101.4 & 105.6 & 136.6 & 210.2 & 279.5 & 1.00 & 1.00 & 1.00 & 0.96 & 0.88 & 0.97\\
        \hline
        \multicolumn{17}{|c|}{Scenario 1: Circle Scenario (agent radius 0.5m)}\\
        \hline
        DMCA & 0.00 & 0.00 & 0.00 & 0.00 & 0.00 & 114.7 & 117.5 & 144.3 & 242.0 & 276.0 & 1.00 & 1.00 & 1.00 & 1.00 & 0.99 & 0.99\\
        \hline
        DMCA-LC & 0.00 & 0.00 & 0.00 & 0.10 & 0.32 & 123.8 & 127.4 & 166.6 &287.0 & NR & 1.00 & 1.00 & 1.00 & 0.90 & 0.63 & 0.91\\
        \hline
        ORCA~\cite{berg2011reciprocal} & 0.00 & 0.00 & 0.00 & 0.00 & 0.02 & 116.1 & 134.2 & 199.1 & 327.5 & NR & 1.00 & 1.00 & 1.00 & 1.00 & 0.75 & 0.95\\
        \hline
        CADRL~\cite{cadrl} & 0.00 & 0.00 & 0.31 & 0.15 & 0.31 & 105.6 & 108.1 & 136.7 & 194.0 & NR & 1.00 & 1.00 & 0.67 & 0.85 & 0.57 & 0.82\\
        \hline
        GA3C~\cite{pedRich} & 0.35 & 1.00 & 1.00 & 1.00 & 1.00 & 104.0 & NR & NR & NR & NR & 0.65 & 0.00 & 0.00 & 0.00 & 0.00 & 0.13\\
        \hline
        BVC~\cite{zhou2017fast} & 0.00 & 0.00 & 0.00 & 0.00 & 0.00 & 213.8 & 372.5 & NC & NC & NC & 1.00 & 1.00 & NC & NC & NC & -\\
        \hline
        BUAVC~\cite{zhu2022decentralized} & 0.00 & 0.00 & 0.00 & 0.00 & 0.00 & 239.3 & 333.1 & NC & NC & NC & 1.00 & 1.00 & NC & NC & NC & -\\
        \hline
        Long et al.~\cite{long} & 0.00 & 0.00 & 0.00 & 0.14 & 0.40 & 102.3 & 106.7 & 139.4 & NR & NR & 1.00 & 1.00 & 1.00 & 0.86 & 0.60 & 0.89\\
        \hline
    \end{tabular}}
    \caption{We tabulate the collision rate, time-to-goal, and success rate for circle scenarios. The maximum simulation time is 500 timesteps and the success rate is computed based on the number of agents reaching the goal before the simulation ends. The results are averaged over 20 trials. The average time-to-goal is computed based on the trials where all robots reach their goal. If no successful trials were observed, we report ``NR'' as time-to-goal because the robots have not reached the final configuration. ``NC'' is used to denote the case where the robots didnt reach their goal in under 500 timestep but did not appear to be deadlocked. We observe that DMCA provides better success rate in the circle scenario with an agent radius of 0.5m.}
    \label{tab:comparision}
\end{table*}

\begin{table*}
    \centering
    \scalebox{0.95}{
    \begin{tabular}{|c|c|c|c|c|c||c|c|c|c|c||c|c|c|c|c|c|}
        \hline
        \multicolumn{17}{|c|}{Scenario 2: Swap Scenario}\\
        \hline
        \multirow{2}{*}{Robots} & \multicolumn{5}{c||}{Collision Rate} & \multicolumn{5}{c||}{Time to Goal} & \multicolumn{5}{c|}{Success Rate (SR)} & \multirow{2}{*}{Avg. SR} \\
        \cline{2-16}
        & 4 & 6 & 10 & 20 & 30 & 4 & 6 & 10 & 20 & 30 & 4 & 6 & 10 & 20 & 30 & \\
        \hline
        DMCA & 0.00 & 0.00 & 0.00 & 0.00 & 0.00 & 123 & 136 & 153 & NR & NR & 1.00 & 1.00 & 1.00 & 0.70 & 0.73 & 0.89\\
        \hline
        DMCA-LC & 0.00 & 0.00 & 0.00 & 0.00 & 0.17 & 122 & 136 & 146 & NR & NR & 1.00 & 1.00 & 1.00 & 0.75 & 0.60 & 0.88\\ 
        \hline
        ORCA~\cite{berg2011reciprocal} & 0.00 & 0.00 & 0.00 & 0.00 & 0.00 & 116 & 161 & 160 & NR & NR & 1.00 & 1.00 & 1.00 & 0.60 & 0.63 & 0.85\\
        \hline
        CADRL~\cite{cadrl} & 0.00 & 0.00 & 0.00 & 0.00 & 0.27 & 106 & 113 & 129 & NR & NR & 1.00 & 1.00 & 1.00 & 0.65 & 0.57 & 0.84\\
        \hline
    \end{tabular}}
    \caption{The collision rate, time-to-goal, and success rate for the swap scenario. The maximum simulation time is 500 steps. In practice, DMCA exhibits the best overall performance.}
    \label{tab:scenario2}
\end{table*}

\begin{figure*}[t]
\centering
  \begin{subfigure}[b]{.17\linewidth}    
        \includegraphics[width=\textwidth, trim={0cm, 0cm, 0cm, 0cm}, clip]{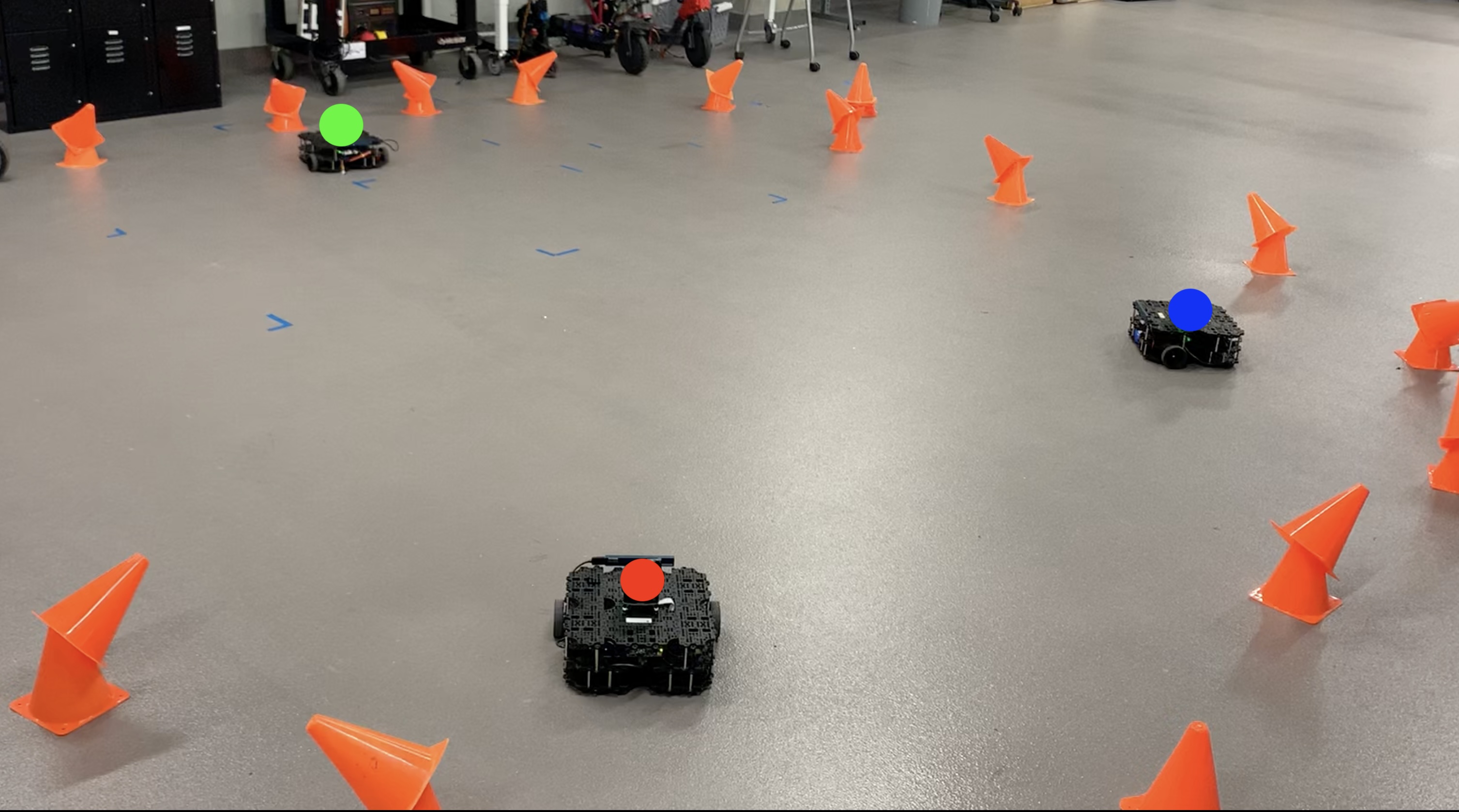}
        \caption{$t=0$}
        \label{img1}
 \end{subfigure}
  \begin{subfigure}[b]{.17\linewidth}    
        \includegraphics[width=\textwidth, trim={0cm, 0cm, 0cm, 0cm}, clip]{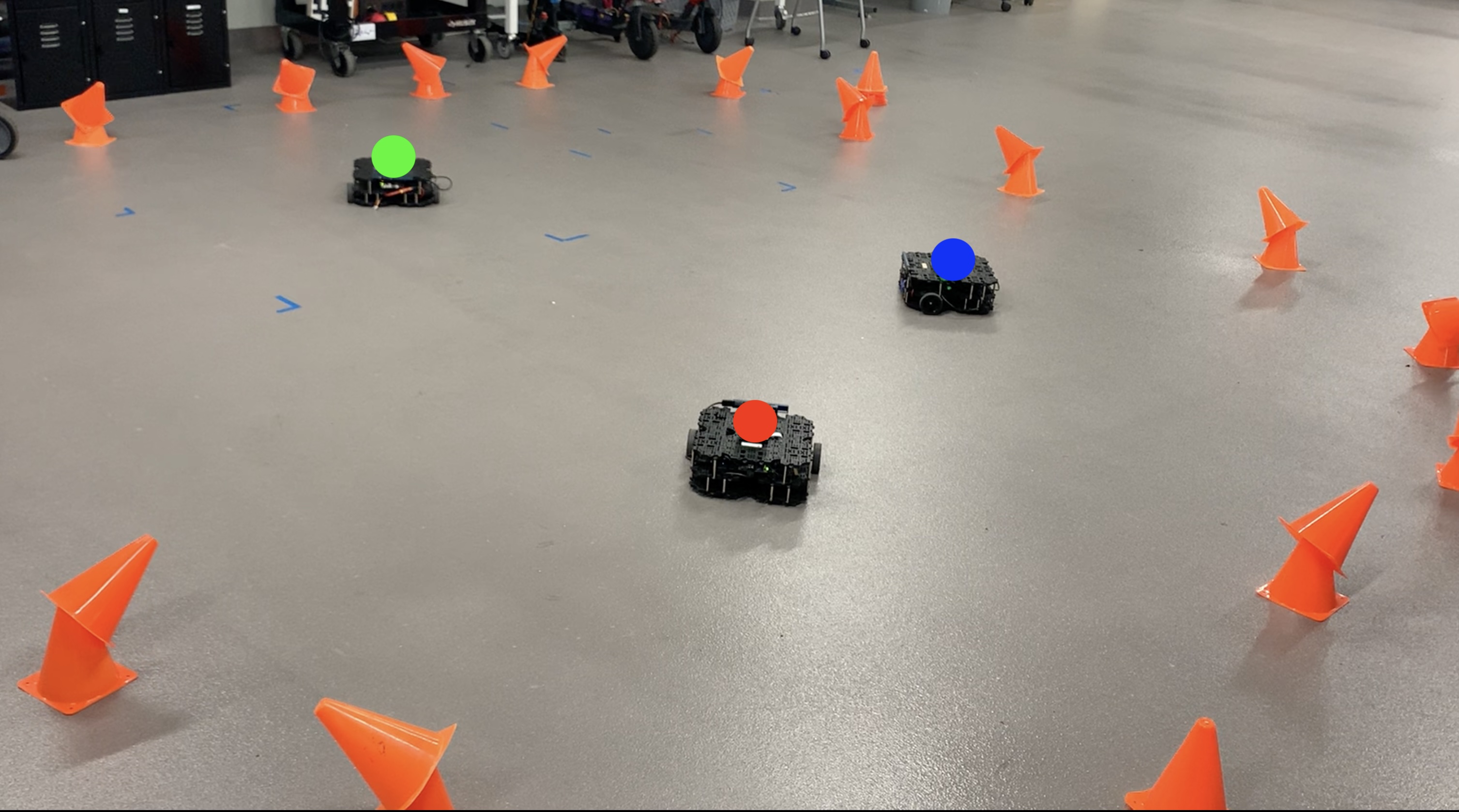}
        \caption{$t=10$}
        \label{img1}
 \end{subfigure}
 \begin{subfigure}[b]{.17\linewidth}  
        \includegraphics[width=\textwidth, trim={0cm, 0cm, 0cm, 0cm}, clip]{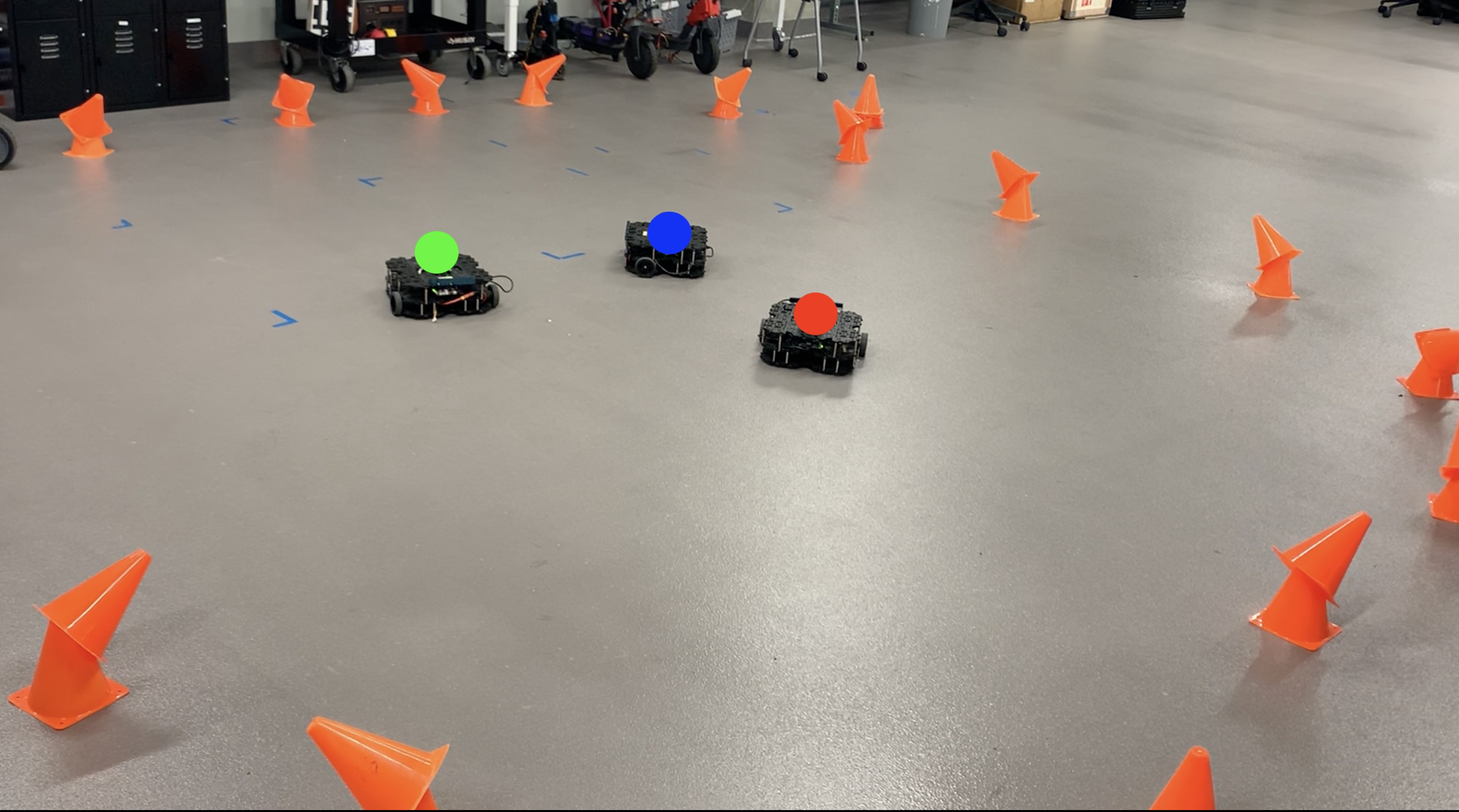}
        \caption{$t=20$}
        \label{img3}
 \end{subfigure} 
  \begin{subfigure}[b]{.17\linewidth}  
        \includegraphics[width=\textwidth, trim={0cm, 0cm, 0cm, 0cm}, clip]{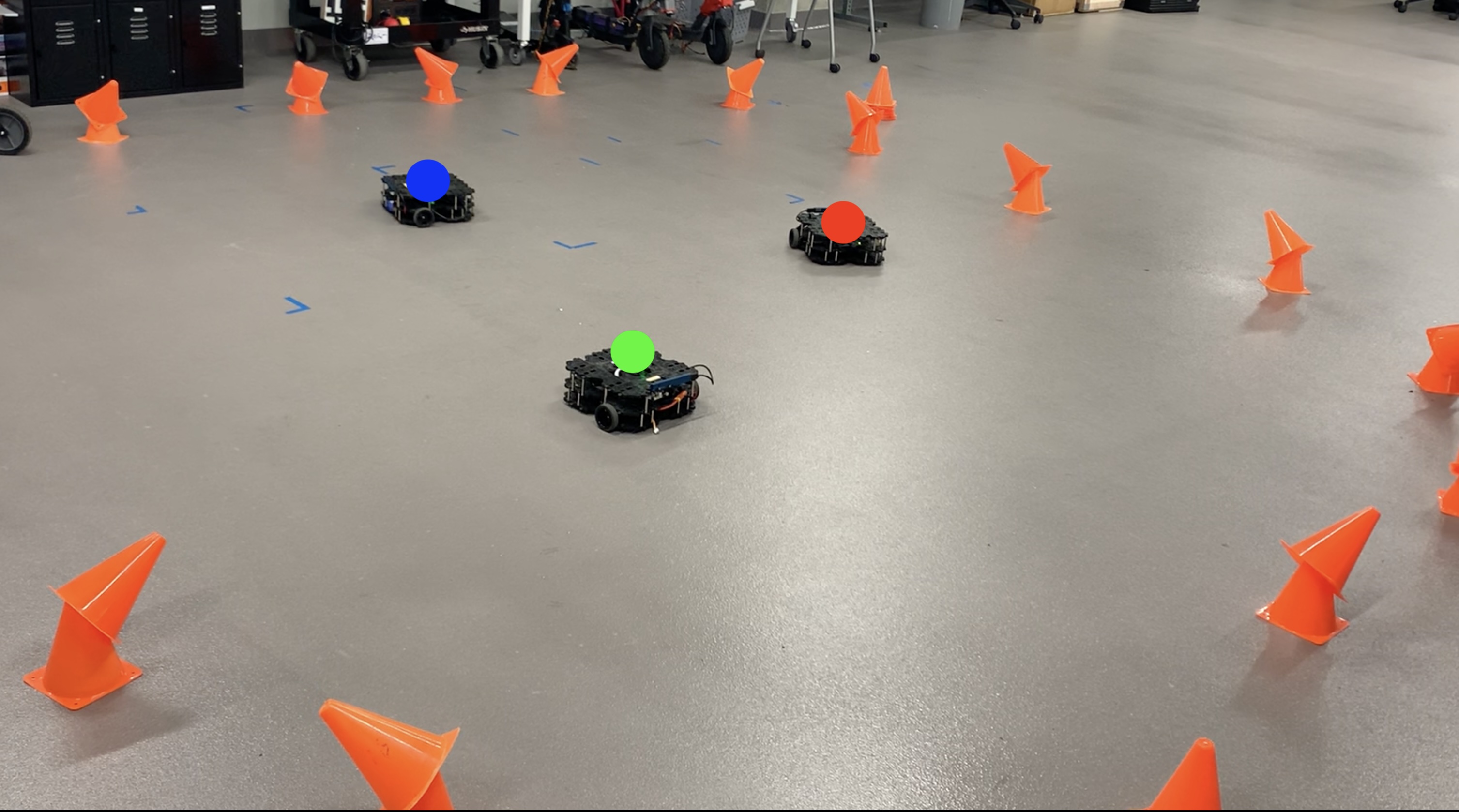}
        \caption{$t=30$}
        \label{img2}
 \end{subfigure}
   \begin{subfigure}[b]{.17\linewidth}  
        \includegraphics[width=\textwidth, trim={0cm, 0cm, 0cm, 0cm}, clip]{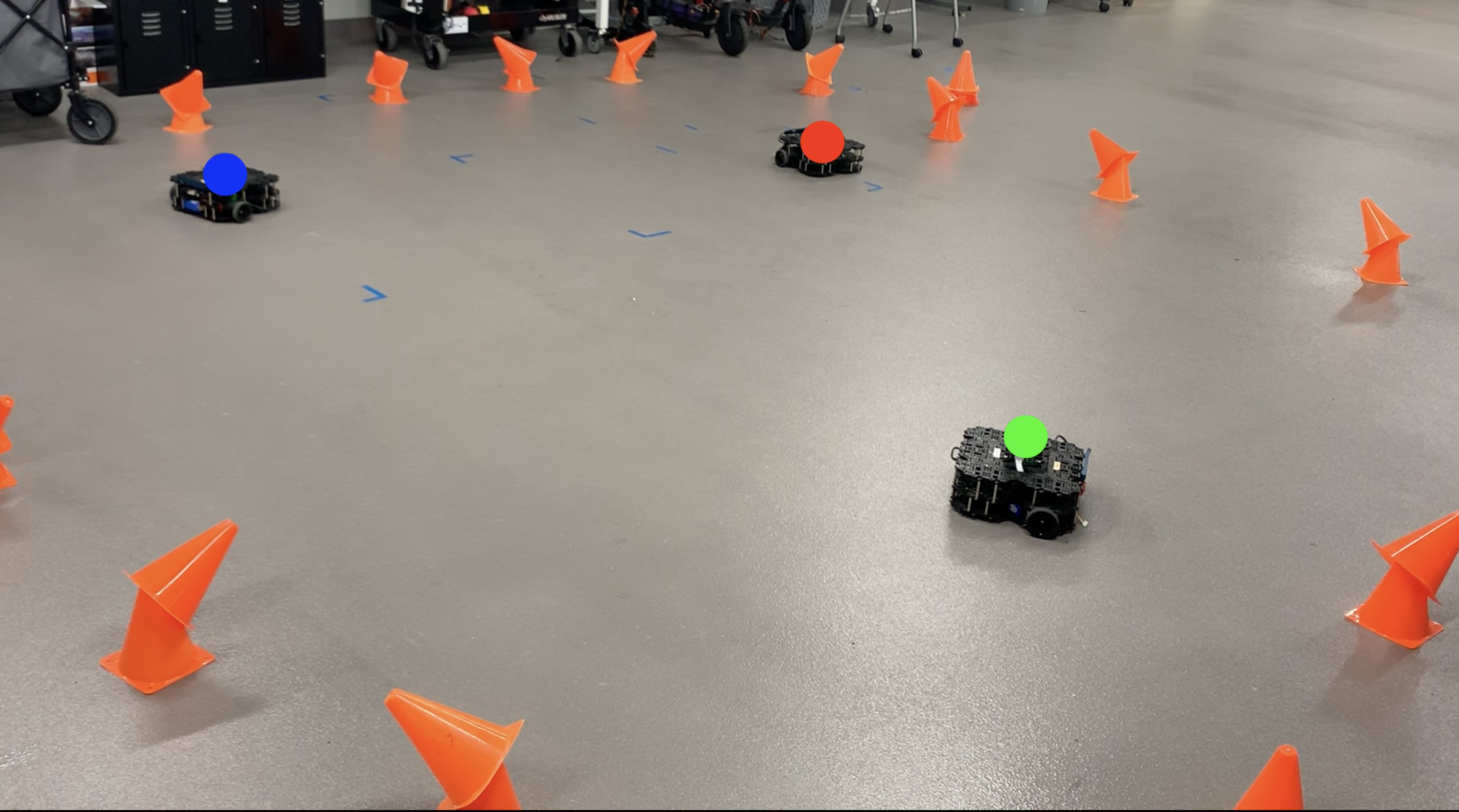}
        \caption{$t=40$}
        \label{img2}
 \end{subfigure}
    \begin{subfigure}[b]{.11\linewidth}  
        \includegraphics[width=\textwidth, trim={1cm, 2cm, 2cm, 1cm}, clip]{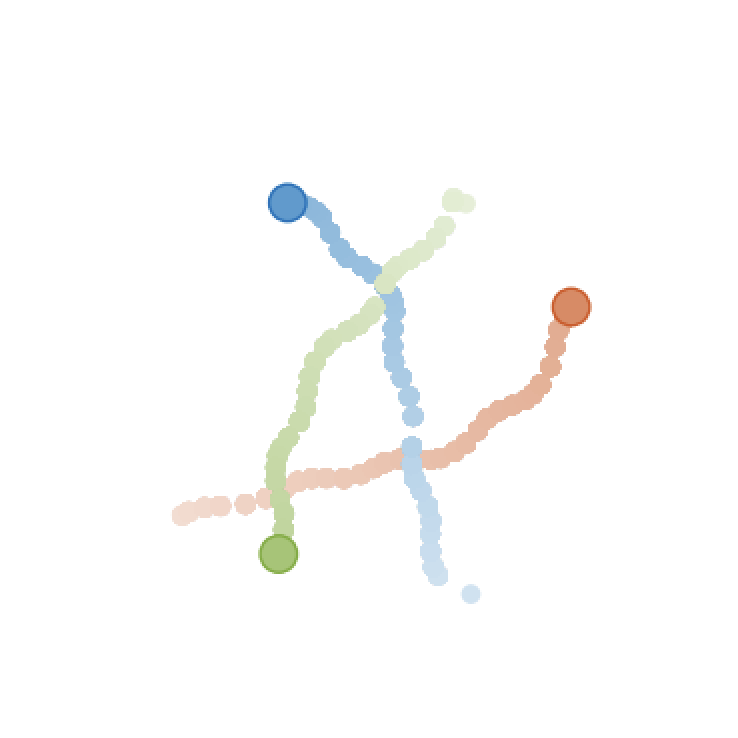}
        \caption{Trajectories}
        \label{img2}
 \end{subfigure}
    \caption{{\color{black}{\bf{Real-world Scenario:}} We evaluate DMCA in a real-world setting with three agents in the circle scenario. Fig 5 (a)-(e) present the snapshot of the execution every 10 seconds. Fig 5(f) illustrates the trajectories followed by each agent. We observe the robots safely navigate to their goal position in this scenario.}}\label{fig:rw}
\end{figure*}
\begin{figure*}[t]
\centering
    \begin{minipage}[t]{0.24\textwidth}
        \centering
        \includegraphics[trim= {50 200 50 0},clip,width=0.99\textwidth]{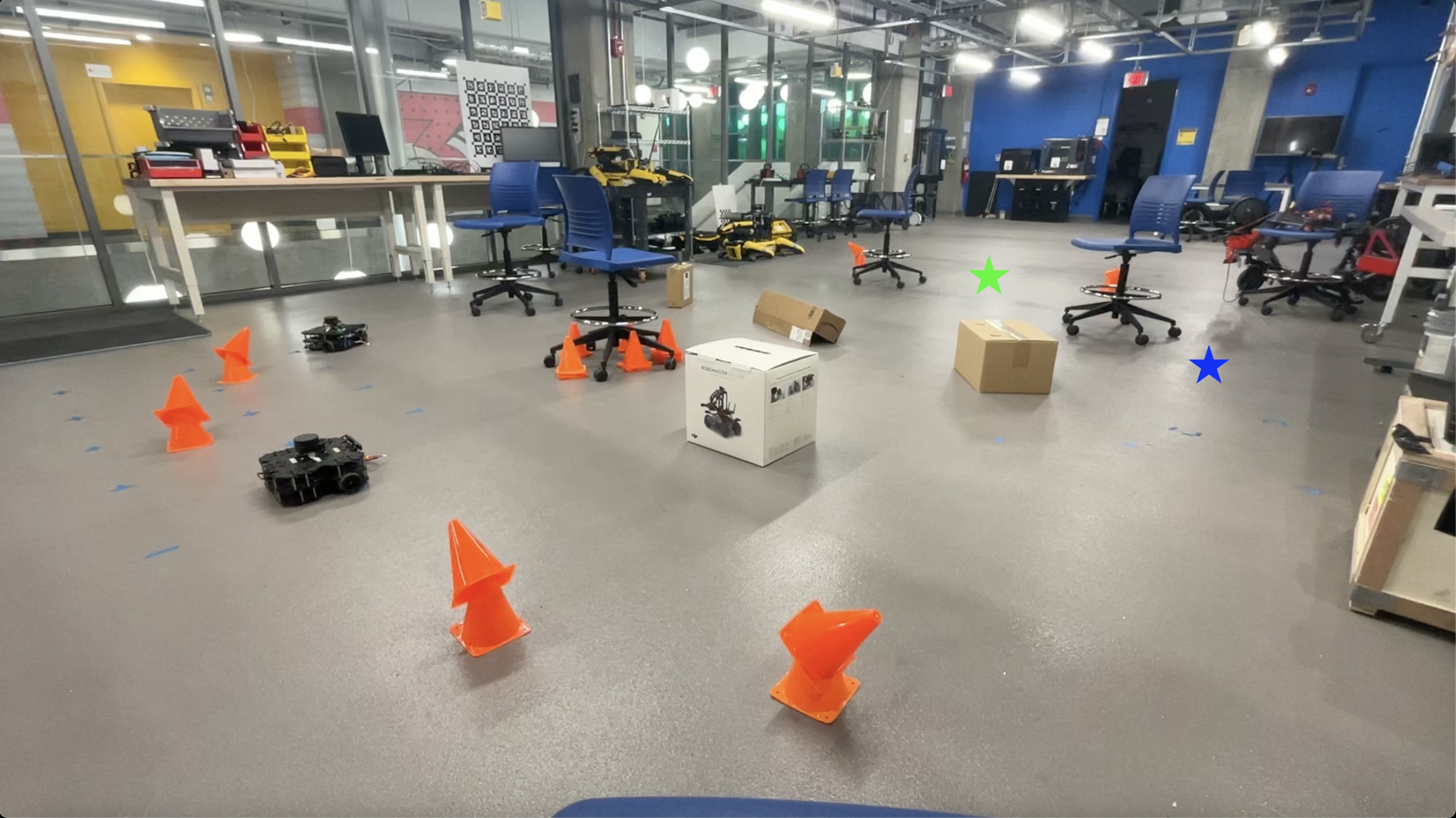}
        \subcaption{t=0s}
    \end{minipage}%
    \begin{minipage}[t]{0.24\textwidth}
        \centering
        \includegraphics[trim= {50 200 50 0},clip,width=0.99\textwidth]{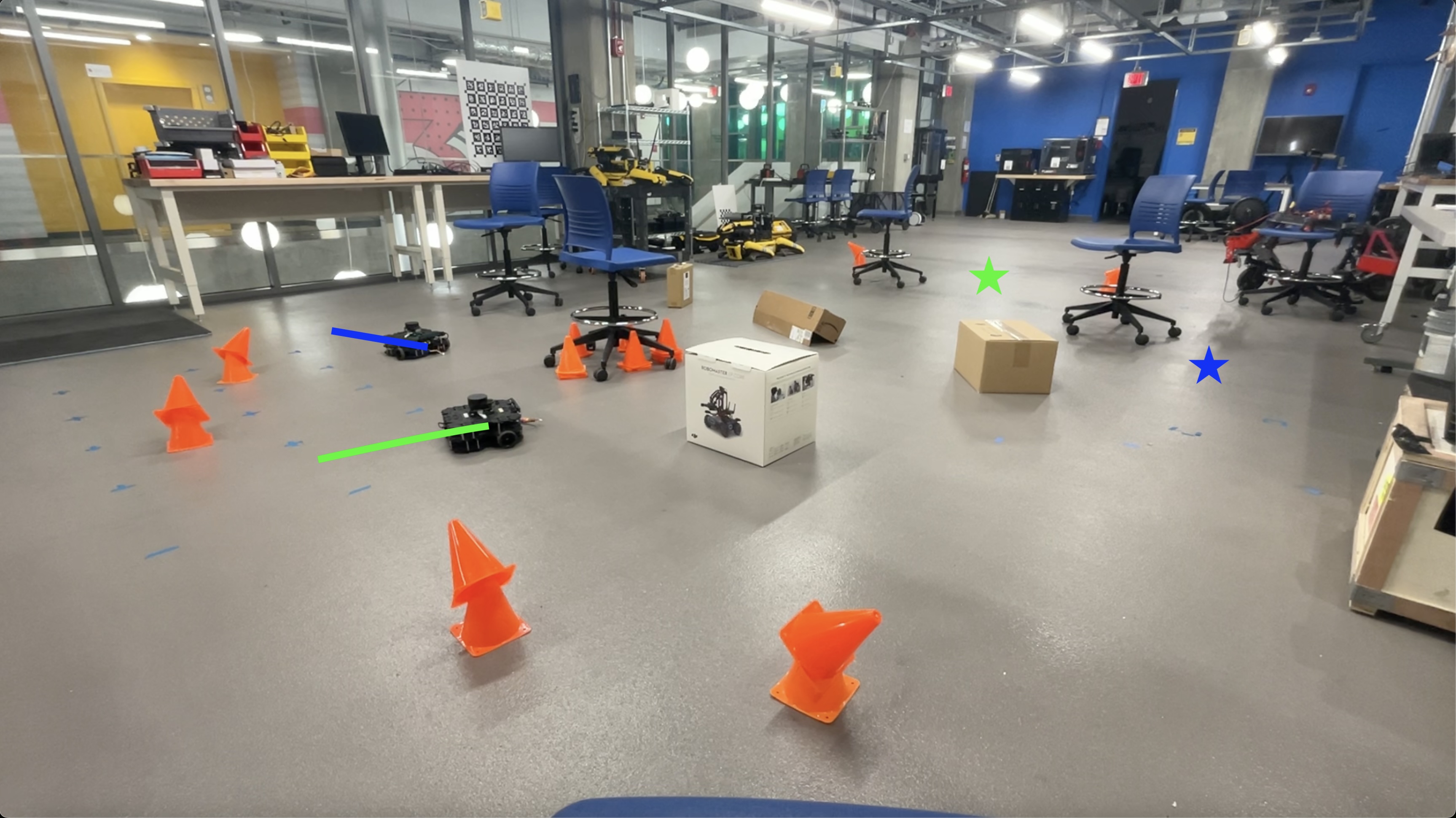}
        \subcaption{t=20s}
    \end{minipage}%
    \begin{minipage}[t]{0.24\textwidth}
        \centering
        \includegraphics[trim= {50 200 50 0},clip,width=0.99\textwidth]{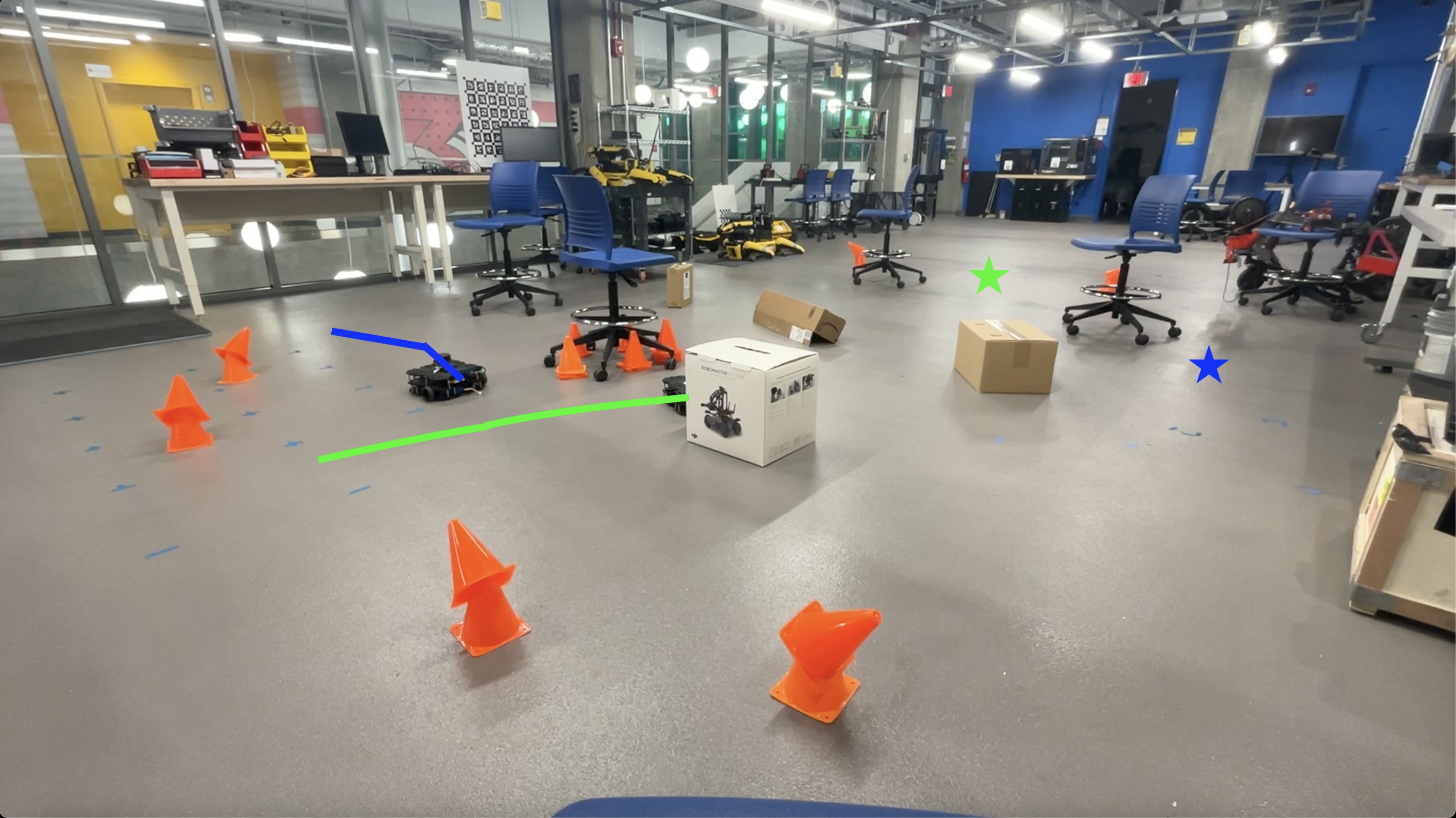}
        \subcaption{t=40s}
    \end{minipage}%
    \begin{minipage}[t]{0.24\textwidth}
        \centering
        \includegraphics[trim= {50 200 50 0},clip,width=0.99\textwidth]{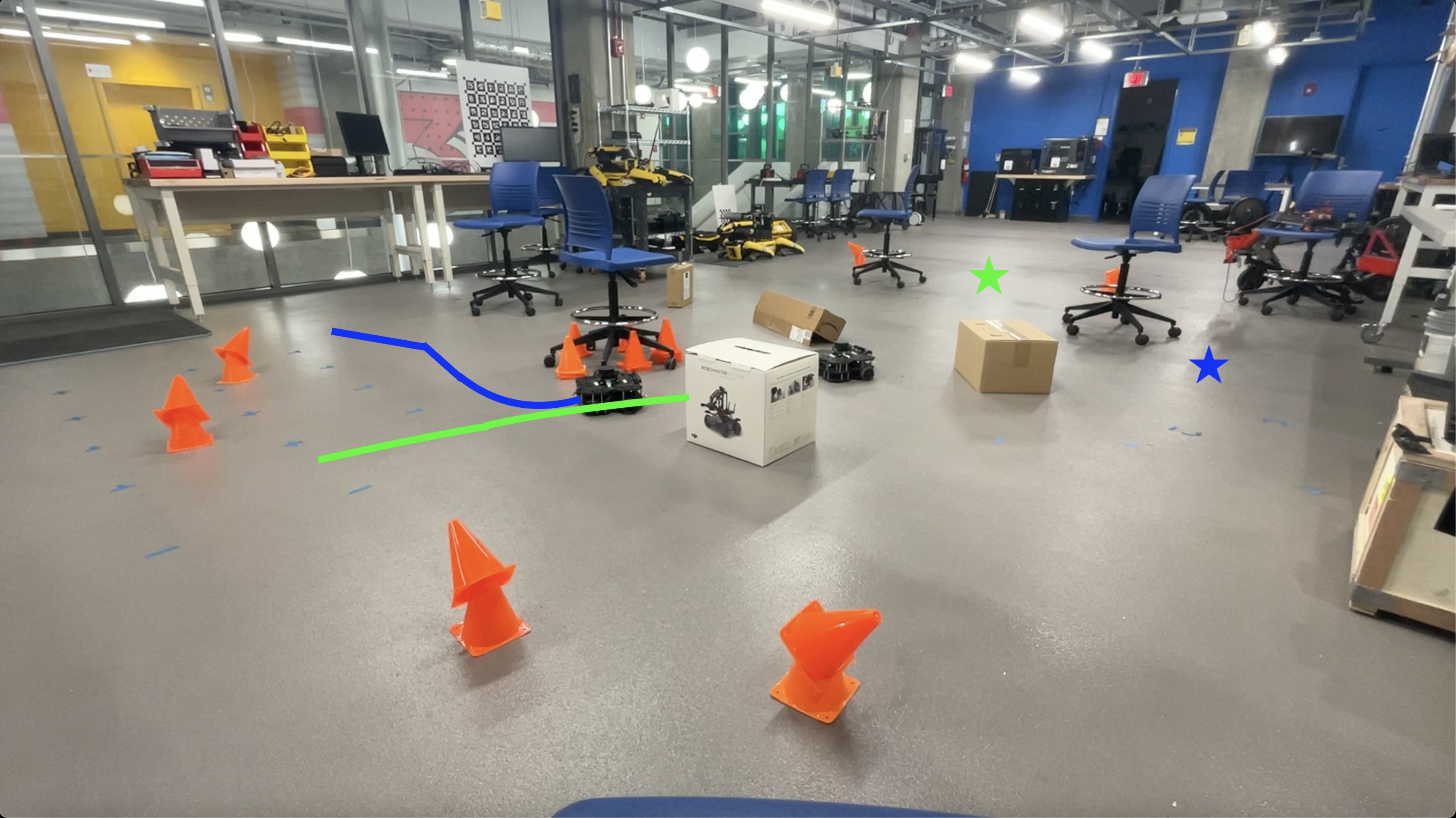}
        \subcaption{t=60s}
    \end{minipage}
    \begin{minipage}[t]{0.24\textwidth}
        \centering
        \includegraphics[trim= {50 200 50 0},clip,width=0.99\textwidth]{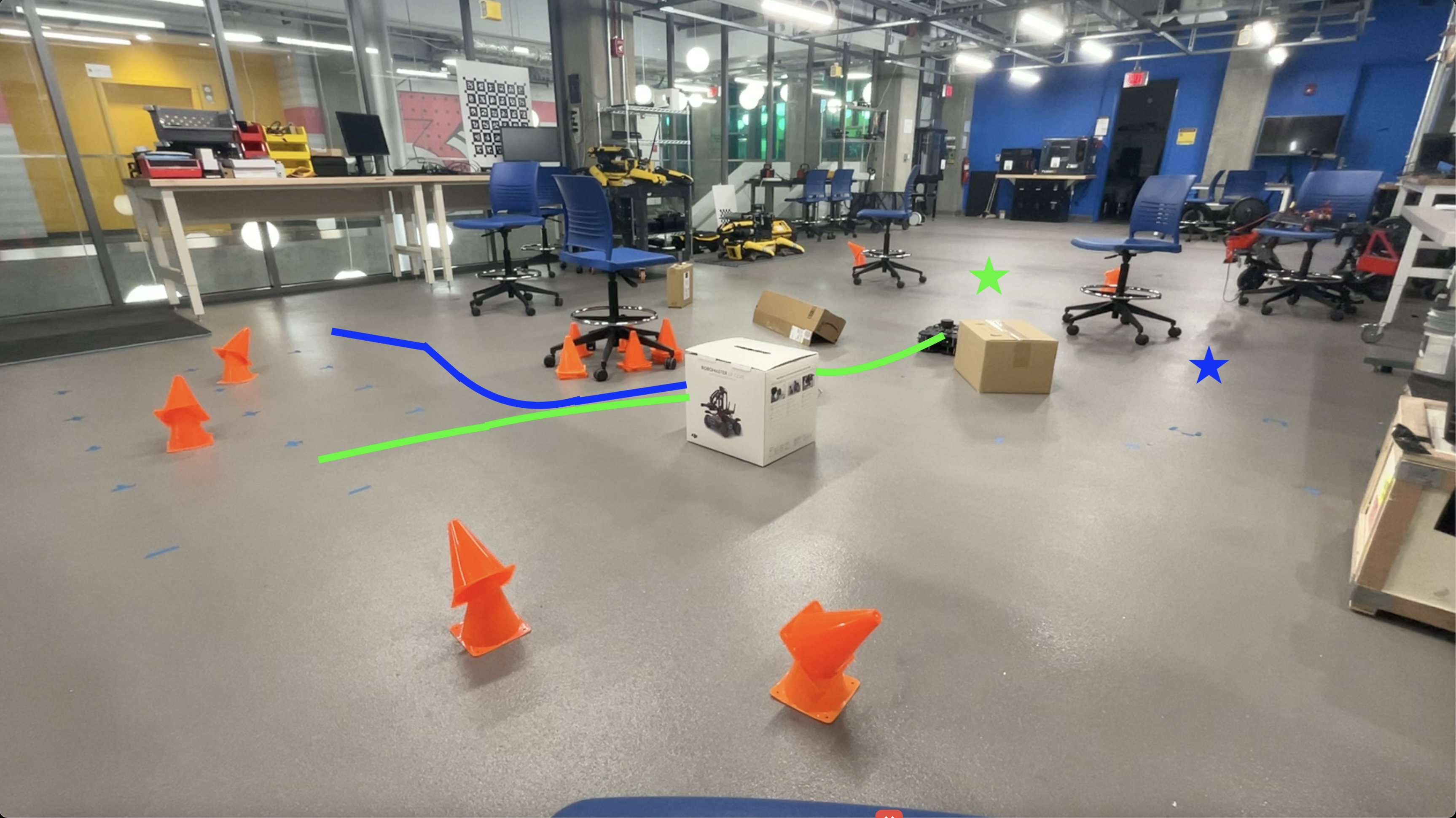}
        \subcaption{t=80s}
    \end{minipage}%
    \begin{minipage}[t]{0.24\textwidth}
        \centering
        \includegraphics[trim= {50 200 50 0},clip,width=0.99\textwidth]{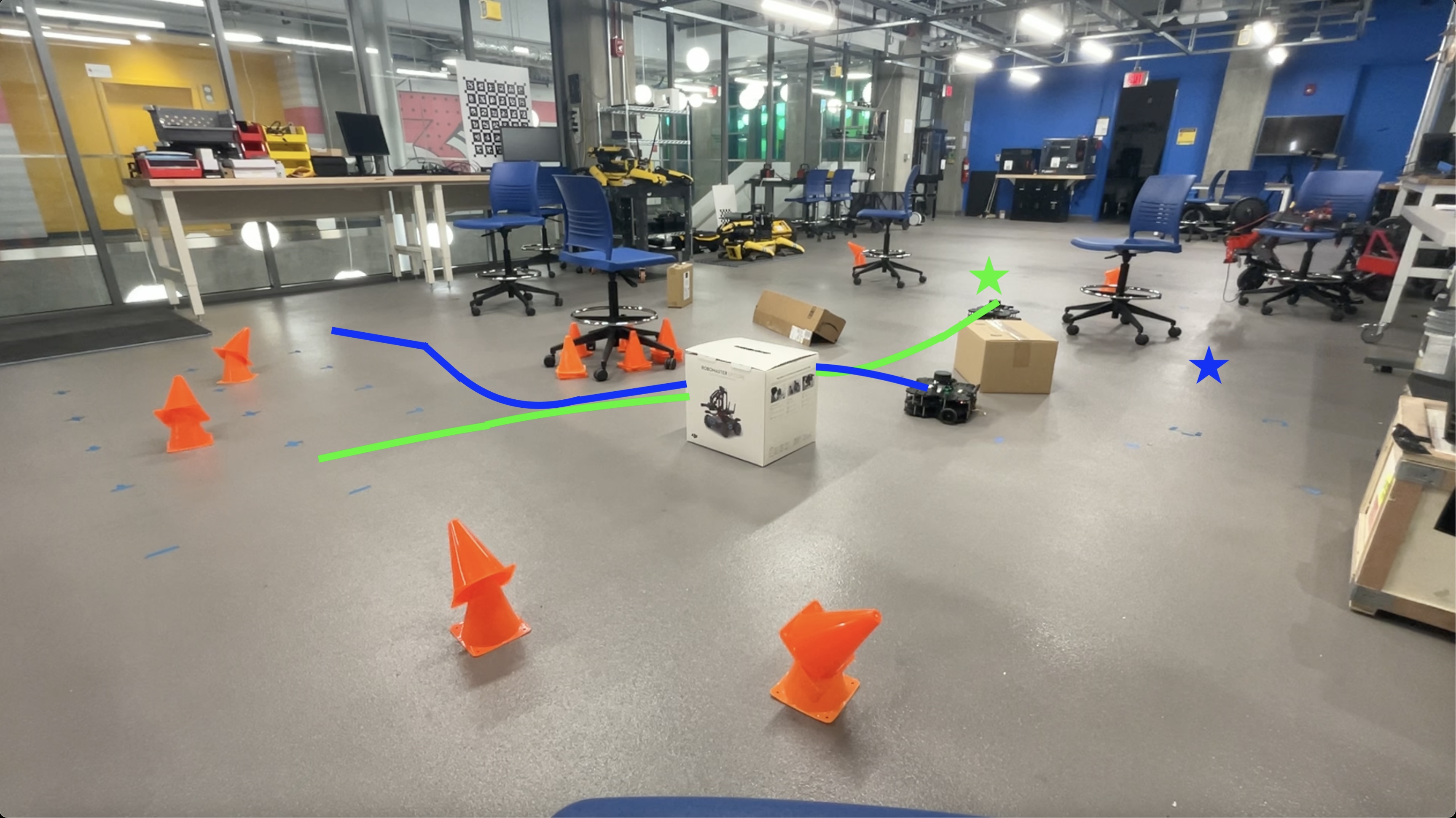}
        \subcaption{t=100s}
    \end{minipage}%
    \begin{minipage}[t]{0.24\textwidth}
        \centering
        \includegraphics[trim= {50 200 50 0},clip,width=0.99\textwidth]{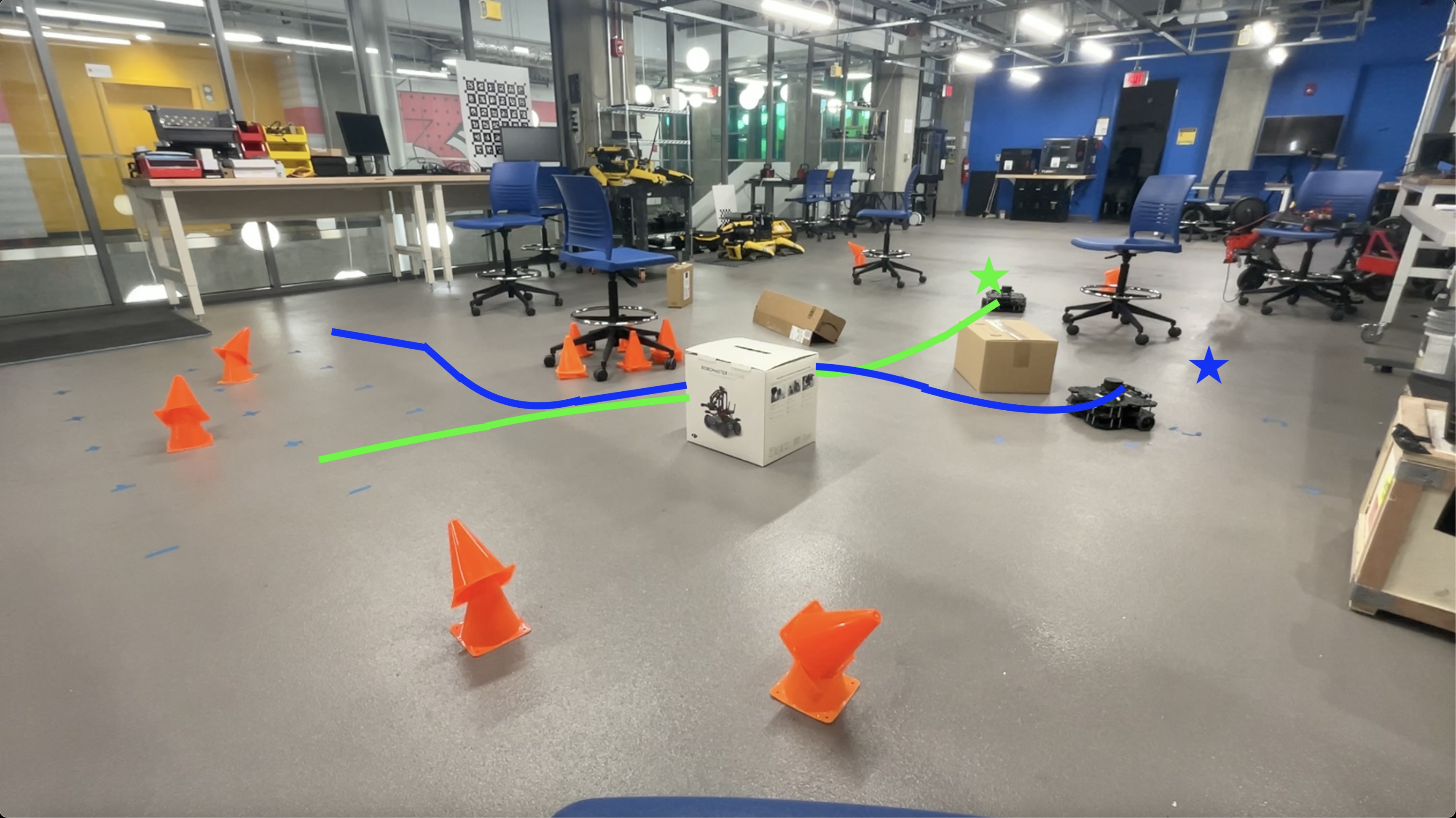}
        \subcaption{t=120s}
    \end{minipage}%
    \begin{minipage}[t]{0.24\textwidth}
        \centering
        \includegraphics[trim= {50 200 50 0},clip,width=0.99\textwidth]{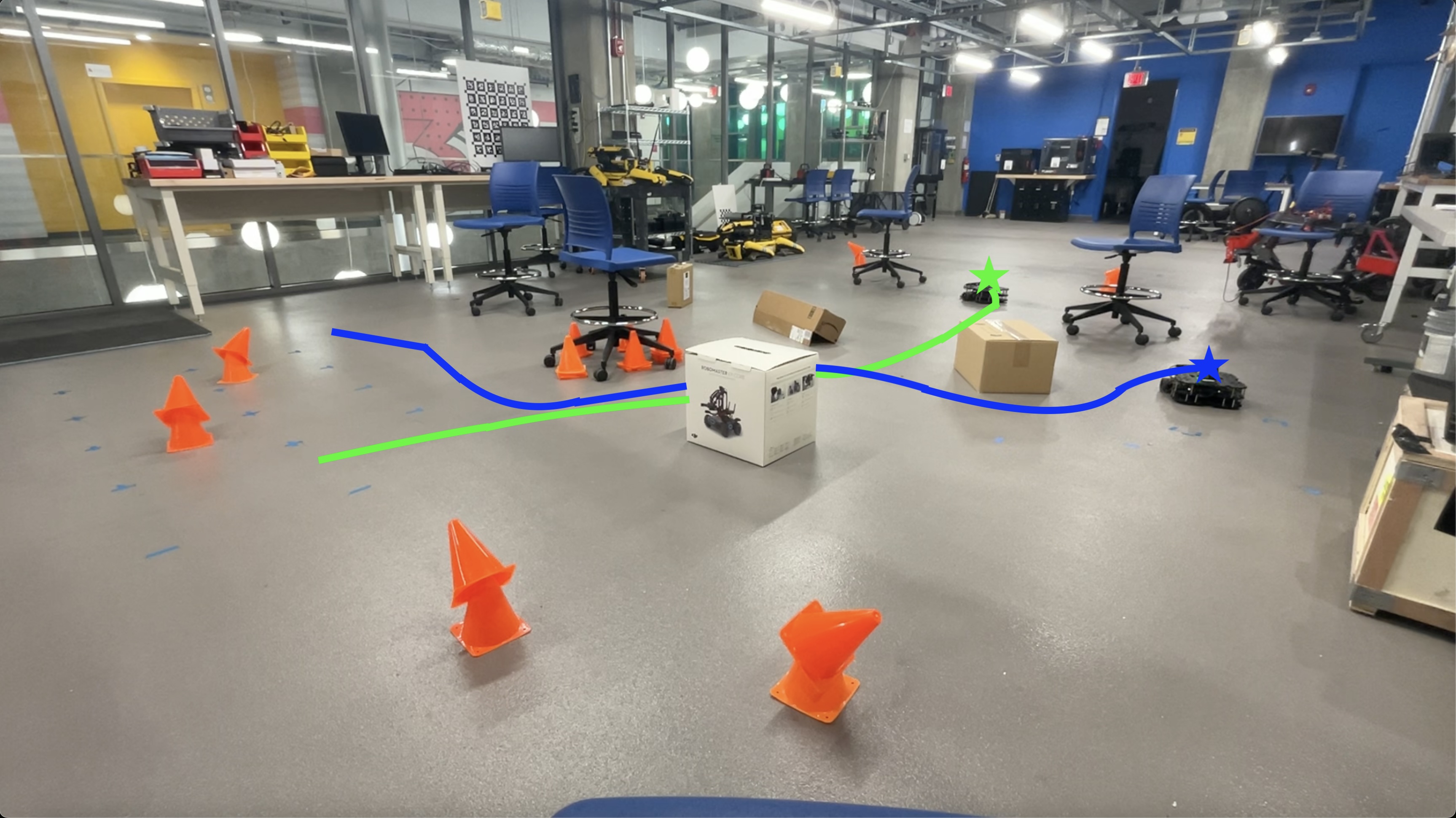}
        \subcaption{t=140s}
    \end{minipage}
    \caption{{\color{black}{\bf{Real-world Scenario:}} We evaluate DMCA in a real-world setting with two agents moving in an environment with multiple static obstacles. Fig (a)-(d) present the snapshot of the execution at regular interval. Our approach navigates the robot successfully with out collisions in this scenario. }}\label{fig:rw_2}
\end{figure*}
\begin{figure}[t]
    \centering
    \includegraphics[width=0.33\textwidth]{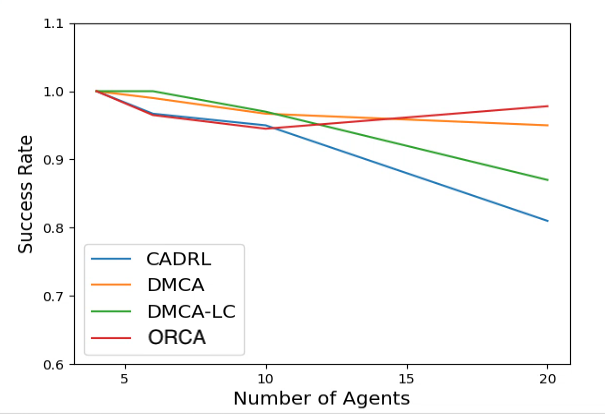}
    \caption{\color{black}\textbf{Success Rate:} Plot illustrates the variation of success rate with number of agents for our proposed method, CADRL~\cite{cadrl}, and ORCA~\cite{berg2011reciprocal}. We observe that our approach (DMCA) results in a good success rate.}
    \label{fig:plot}
\end{figure}
\begin{figure}[h!]
   \centering
            \begin{subfigure}[b]{.49\linewidth} 
            \centering
                \includegraphics[width=0.99\linewidth]{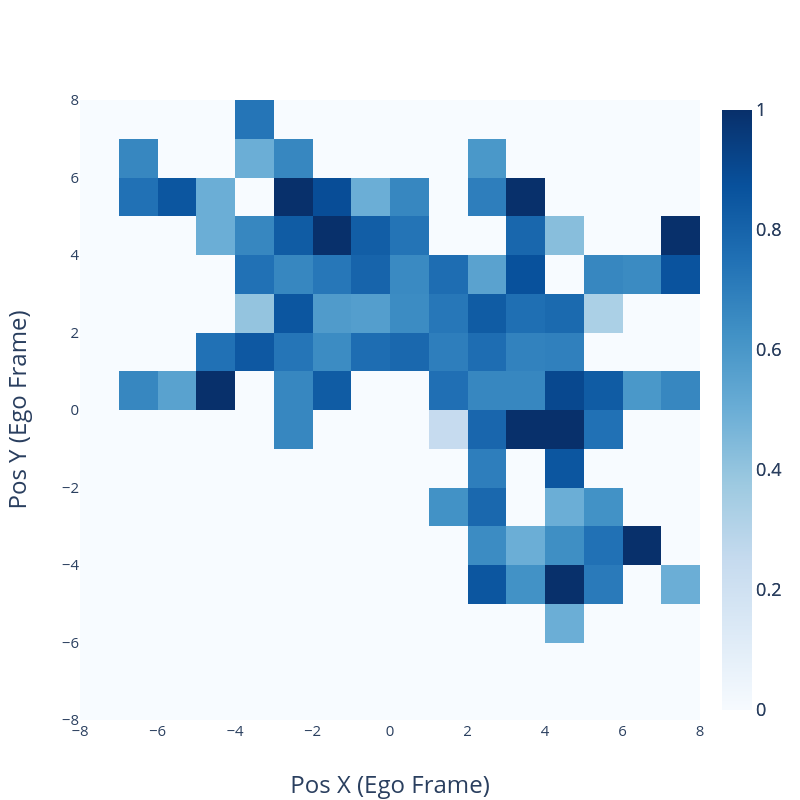}
                \caption{DMCA}
            \end{subfigure}
            \begin{subfigure}[b]{.49\linewidth}  
            \centering
                \includegraphics[width=0.99\linewidth]{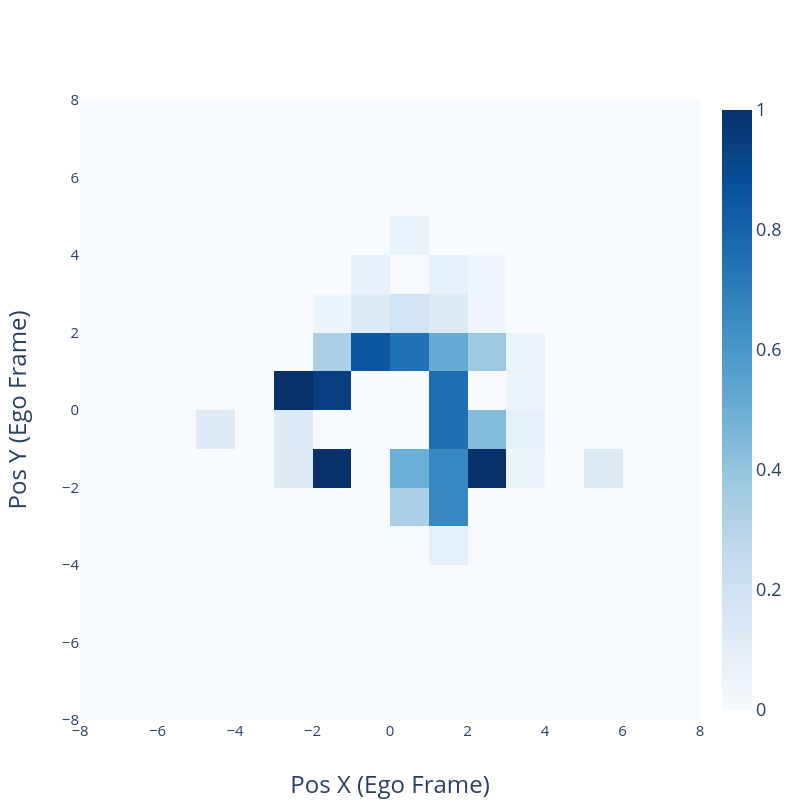}
                \caption{DMCA-LC}
            \end{subfigure} 
            \caption{The figure shows a 2D histogram representing the position of neighboring agents in the robot's ego frame. The histogram bins show the average value for the communication link. The Darkest Blue (value of 1) represents a communication link, while the lightest blue (value of 0) represents a lack of communication link. We observe {\color{black}DMCA-LC} communicates with closer neighbors in the ego frame compared to {\color{black}DMCA}.}
            \label{fig:hist}
\end{figure}
\begin{figure}[h!]
    \centering
    \begin{subfigure}[b]{.49\linewidth}    
    \centering
    \includegraphics[width=0.9\linewidth, trim={0cm, 1cm, 1cm, 0cm}, clip]{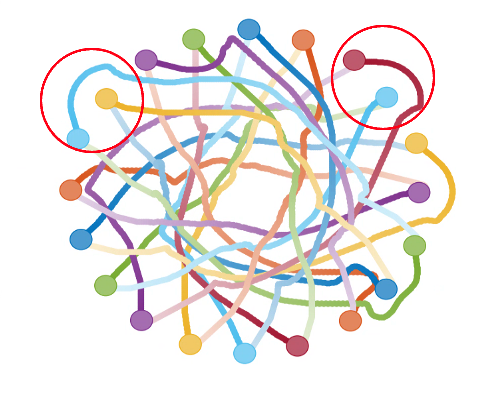}
        \caption{DMCA}
    \end{subfigure}
    \begin{subfigure}[b]{.49\linewidth}  
    \centering
        \includegraphics[width=0.94\linewidth, trim={0cm, 1cm, 1.5cm, 0cm}, clip]{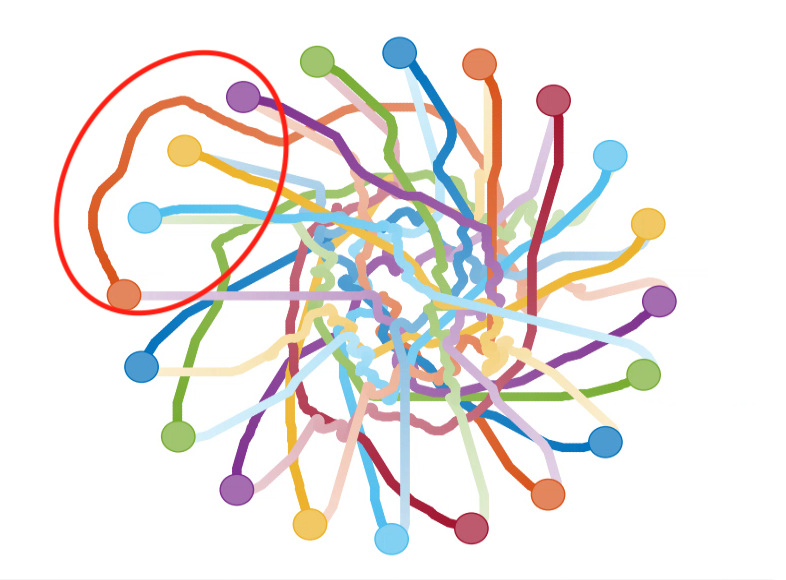}
        \caption{DMCA-LC}
    \end{subfigure} 
    \caption{ {\color{black}DMCA through the concurrent learning of navigation and communication appears to have learnt to use the communicated goal information from neighbors and navigate the robot around neighbors' goals.}}
    \label{fig:goal-aversion}
\end{figure}

\subsection{Computational Setup}~\label{sec:compsetup}
Our method was implemented on an Intel Xeon 4208 CPU with 32 GB RAM and an Nvidia GeForce RTX 2080 Ti graphic card. We use TensorFlow and Python for the deep learning implementation. We use gym-collision avoidance and a GA3C-CADRL package~\cite{Everett18_IROS} to implement our method and for our evaluations. The BVC and BUAVC implementations are from~\cite{zhu2022decentralized}.

We train our model in a curriculum setup. Initially, we use a set of simple scenarios with upto four agents to train the network weights. Subsequently, in phase 2, the scenario complexity is raised by increasing the agents in the training environments. The training process took $\sim2$ days to converge on our setup.

\subsection{Baselines}
We compare our method against prior model- and learning-based decentralized methods. {\color{black}We choose ORCA~\cite{berg2011reciprocal}, CADRL~\cite{cadrl}, GA3C-CADRL~\cite{pedRich}, BVC ~\cite{zhou2017fast}, BUAVC~\cite{zhu2022decentralized}, and Long et al.~\cite{long} as our baselines for comparison. 
We present two versions of our algorithm: 1) DMCA: trained without the negative reward for the communication link, that is $\lambda_{comm} = 0$ and 2) DMCA-LC: trained with a negative reward for the communication link, $\lambda_{comm} = 0.0001$ in our evaluations. Both DMCA and DMCA-LC communicate considerably lower than broadcast, while DMCA-LC communicates more conservatively.}

\subsection{Trajectories}


\begin{table*}[h!]
    \centering
    \scalebox{0.9}{
    \begin{tabular}{|c|c|c|c|c|c|c|c|}
    \hline
        Agents & DMCA & DMCA-LC &  Broadcast (within 3m) & Broadcast (within 5m) & Broadcast (within 6m) & Broadcast (within 8m) & Broadcast (all agents)\\
    \hline
        2 &  76 &  1 &   0 &   42 &  57 &  82 & 115\\
        4 & 234 & 13 &  30 &  200 & 247 & 315 & 345 \\
        6 & 252 & 59 & 123 &  360 & 418 & 532 & 590 \\
    \hline
    \end{tabular}}
    \caption{{\color{black}Number of communication links made by a single agent in circle scenario. DMCA-LC significantly reduced the number of pairwise communications. The broadcast communication values are based on the number of agents that are within the said distance bound with the ego agent. }}
    \label{tab:comm_link}
\end{table*}

In Figure~\ref{fig:traj}, we compare the generated trajectories between our proposed methods DMCA and DMCA-LC with CADRL, GA3C-CADRL, and ORCA. In this scenario, we observed DMCA to produce smoother trajectories than DMCA-LC. In addition, the other learning methods result in collisions. {\color{black}In Figures~\ref{fig:traj_form}, we compare the navigation performance in a formation scenario. This formation is particularly challenging owing to the constrained initial configuratinon.
DMCA results in the best success rate, followed by DMCA-LC, and CADRL. ORCA resulted in a success rate of zero. The formation scenario was not used for training, hence DMCA's performance shows it generalizes to other scenarios.

In Figure~\ref{fig:traj_static_2}, we show the trajectories generated by DMCA in a benchmark with static obstacles. DMCA navigated the robots safely in these scenarios.}

\subsection{Collision Rate, Success Rate, and Time-to-Goal}
We compare our proposed method with other baselines in terms of the number of collisions, time-to-goal, and the success rate in reaching the goal formation. We define the success rate as the fraction of the total number of agents that reach the goal without colliding or getting deadlocked. 

We consider two scenarios: 1) A circle scenario with robots placed on the perimeter of a circle and 2) A swap scenario where the robots arranged into two columns swap their position with adjacent robots in the other column. Tables~\ref{tab:comparision} and~\ref{tab:scenario2} summarize our observations.

{\bf{Scenario 1 (Circle):}}
In the circle scenario, we observe that our proposed method results in the fewest collisions compared to the other methods. With regards to the time-to-goal, CADRL results in a shorter time-to-goal in the scenarios where the agents remained collision-free. {\color{black}Following DMCA, ORCA and DMCA-LC have the higher success rate in this scenario.}

{\bf{Scenario 2 (Swap):}}
We observed DMCA resulted in fewer collisions and a higher success rate in our method than in other baselines. The time-to-goal values are higher than CADRL. {\color{black}Averaging the success rate computed over the cases with 4 to 30 robots, DMCA performs the best, followed by DMCA-LC, ORCA, and CADRL.} 

\subsection{Random Scenario}
In this evaluation, we consider a scenario with random start and end points with agents of random radii. In Figure~\ref{fig:plot}, we plot the success rate (averaged over 50 trials) with respect to the number of agents. We observe DMCA results in a higher success rate than other learning-based method.

\subsection{Comparision with Zhai et al (SelComm~\cite{selcomm})}

We compare our approach with a relevant prior work (SelComm~\cite{selcomm}) for comparison. 
Despite the absence of publicly available training checkpoints for the network, we have made our best effort to recreate the benchmarks for the \em{Group Swap} and \em{Random scenario} as outlined in their study. We acknowledge that variations in implementation details could influence the outcomes. The metric values for our method in these benchmarks are appended to the Table~\ref{tab:selcomm_random} presented in SelComm~\cite{selcomm} and included in this manuscript. 

First, we compare our approach in the random scenario benchmark used on~\cite{selcomm}. As in the paper, we consider a circular region of $10m$ radius where static obstacles are randomly place. The start and goal position for each robot is randomly generated such that the start and the goal position are $8-10$ meters apart as in~\cite{selcomm}. We approximate irregularly space static obstacles as circular disk or collection of circular disk for our approach. 

In table~\ref{tab:selcomm_random} we append the results obtained with our proposed approach (DMCA) in terms of the success rate (SR), collision rate (CR), and average Time to goal (NT) for a team of $8$, $16$, and $24$ robots to the Table 2 in~\cite{selcomm}. We would like to highlight that our approach results in $100\%$ success in these scenarios while maintaining a similar time-to-goal performance.

\begin{figure*}[h!]
    \centering
    \begin{minipage}[t]{0.3\linewidth}
            \centering
            \includegraphics[width=\textwidth]{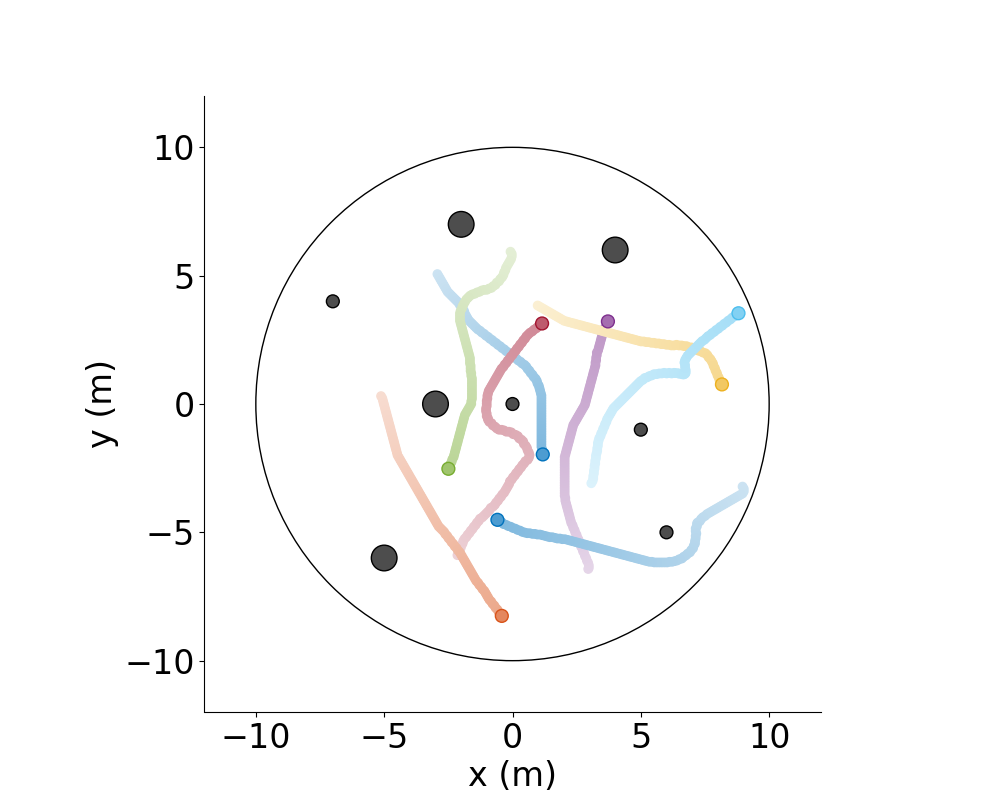}
            \subcaption{8 Agents}
            \label{fig:selcomm_groupswap_img}
        \end{minipage}
        \begin{minipage}[t]{0.3\linewidth}
            \centering
            \includegraphics[width=\textwidth]{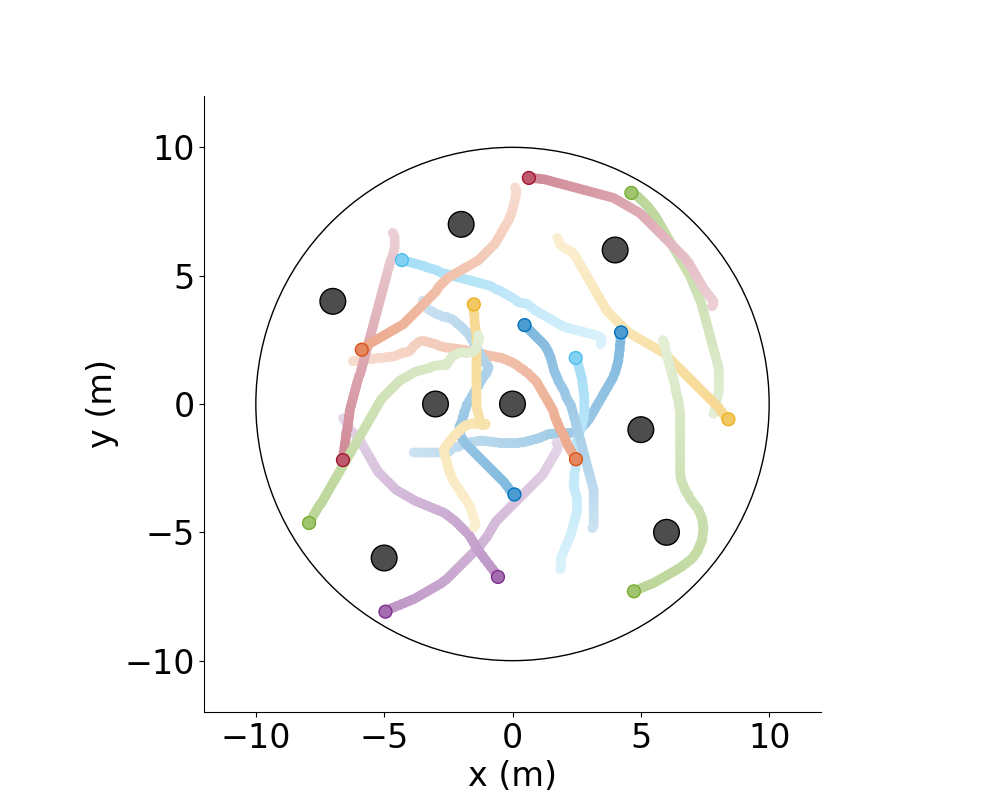}
            \subcaption{16 Agents}
            \label{fig:selcomm_groupswap_plot}
        \end{minipage}
        \begin{minipage}[t]{0.3\linewidth}
            \centering
            \includegraphics[width=\textwidth]{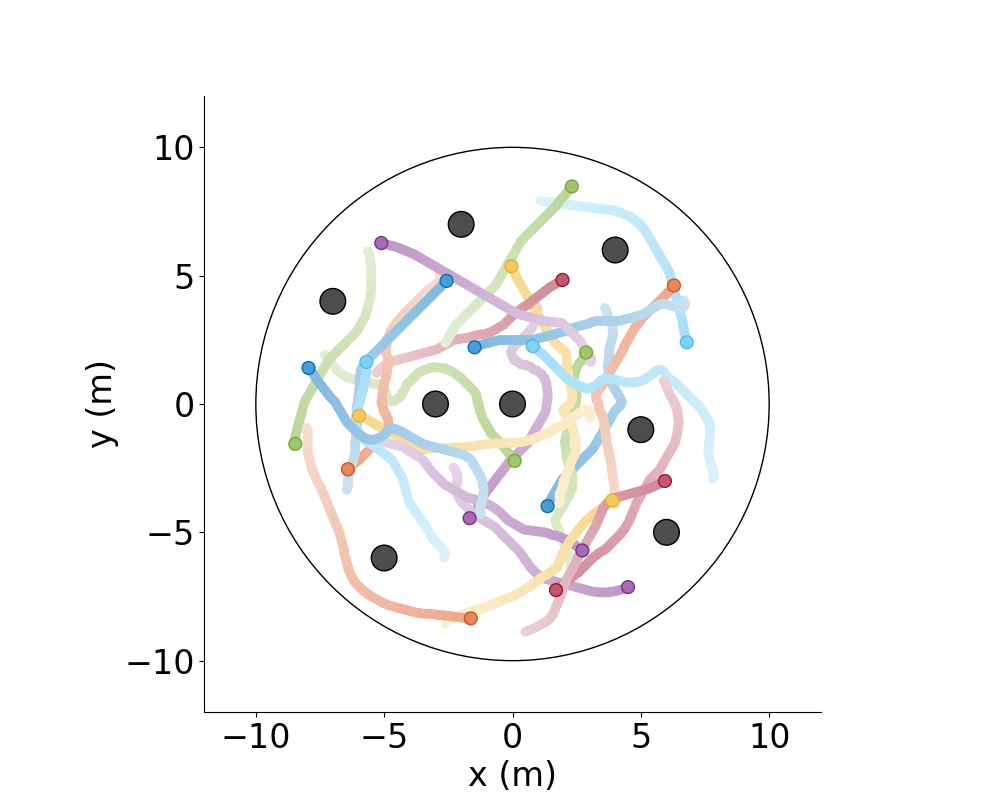}
            \subcaption{32 Agents}
            \label{fig:selcomm_groupswap_plot}
        \end{minipage}
        \caption{\textbf{Random Scenario: }The environment is a circular region with a 10-meter radius, containing randomly distributed static obstacles and multiple robots. Each robot starts from a random position and moves toward a randomly generated goal. The distance from the start position to the goal position ranges between 8 and 10 meters. The figure (a)-(c) illustrates 8, 16, and 24 robots navigating using our proposed method DMCA.}
\end{figure*}

\begin{table*}[h!]
    \centering
    \scalebox{0.95}{
    \begin{tabular}{c|c|c|c|c|c|c|c|c|c|}
            \hline
             & \multicolumn{3}{|c|}{N = 8} & \multicolumn{3}{|c|}{N = 16} & \multicolumn{3}{|c|}{N = 24} \\
             \hline
             Method & SR & CR & NT & SR & CR & NT & SR & CR & NT\\
             
             Sensor-level~\cite{long} & 27.88 & 23.50 & 138 & 23.94 & 35.05 & 134 & 11.87 & 52.59 & 132 \\
             TarMAC~\cite{Tarmac} & 61.10 & 22.20 & 108 & 53.07 & 43.56 & 97 & 44.37 & 50.65 & 95 \\
             ATOC (DVN)~\cite{jiang2018learning} & 72.32 & 19.20 & 101 & 42.10 & 44.19 & 97 & 26.35 & 56.65 & 96 \\
             SelComm (Distance)~\cite{selcomm} & 65.54 & 17.11 & 102 & 62.40 & 35.61 & 93 & 34.51 & 50.75 & 101 \\
             SelComm (DVN)~\cite{selcomm} & 76.14 & 17.59 & 95 & 58.43 & 34.46 & \textbf{87} & 43.67 & 47.42 & \textbf{93} \\
             DMCA (Ours) & \textbf{100.00} & \textbf{0.00} & \textbf{94.48} & \textbf{100.00} & \textbf{0.00} & 96.43 & \textbf{100.00} & \textbf{0.00} & 98.42\\
             \hline
        \end{tabular}}
        \caption{In our evaluation, we consider random scenarios as described in SelComm~\cite{selcomm}. To their existing results table, we append only the performance metrics of our approach. Our method is highlighted for achieving the highest success rates, while maintaining comparable time-to-goal metrics in these scenarios.}
        \label{tab:selcomm_random}
\end{table*}

\begin{figure}[h!]
    \centering
    \begin{minipage}[t]{0.45\textwidth}
            \centering
            \includegraphics[width=\textwidth]{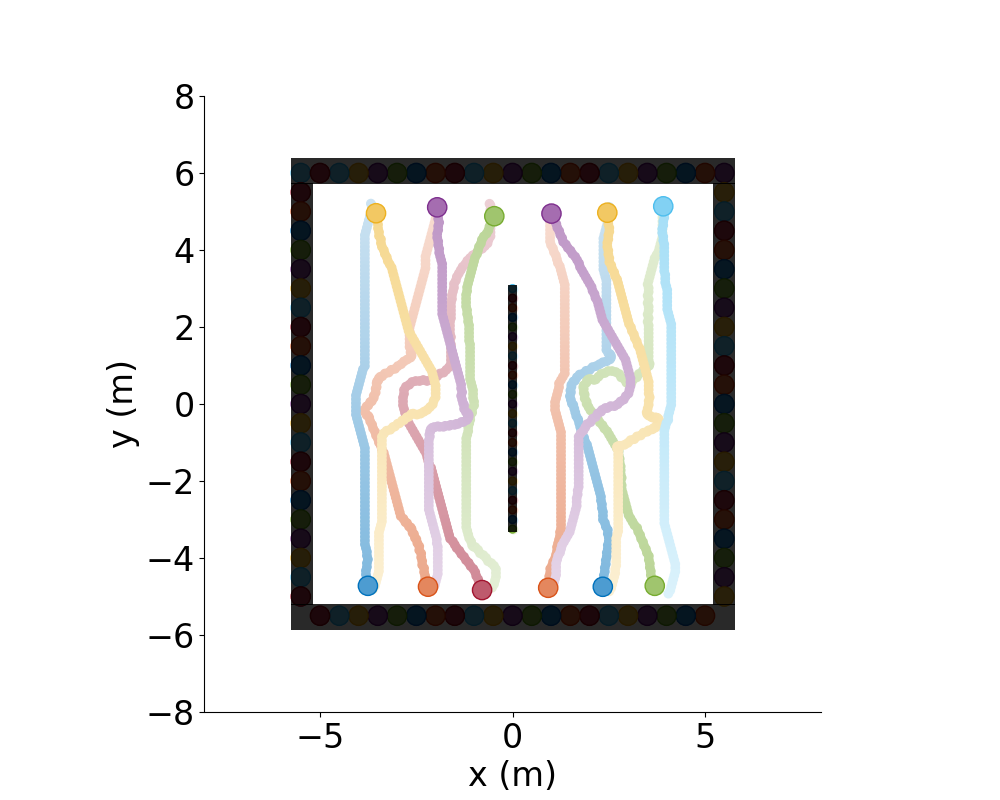}
            \caption{Trajectories of 12 robots performing group swap using our proposed approach (DMCA).}
            \label{fig:selcomm_groupswap_img}
        \end{minipage}\quad
\end{figure}
Second, we compare our approach in the group swap scenario where the robots are divided into two groups (each group has six robots) moving in opposite directions to swap their positions. A slender obstacle separates the passageway. We illustrate the trajectories generated by our approach in Fig.~\ref{fig:selcomm_groupswap_img}. 

{\color{black}\subsection{Real-world Evaluations}
We evaluate DMCA in a real-world setting with robots in a circle scenario. In Fig.~\ref{fig:rw}, we present snapshot of the experiments and the trajectories generated by DMCA in this setting. The agent's self-localize using AMCL\footnote{http://wiki.ros.org/amcl} and hence their positions are noisy. The sensor data is communicated to a computer running DMCA and the velocity commands are communicated back to the robot and the communications are implemented using ROS topics\footnote{https://www.ros.org}.}

\subsection{Discussion}
We compare DMCA and DMCA-LC to show the effect of using a negative reward for communication. In Table~\ref{tab:comm_link}, we tabulate the number of communications a single agent makes in circle scenario. In this scenario, an agent can communicate with a maximum of 3 agents at any time step. We observe that DMCA-LC uses significantly fewer communications than DMCA. {\color{black}We estimate the communication links per agent with broadcast as the product of time step to goal (from Table 2) and number of neighbors (robots other than the ego agent in the environment). From Table 4, we observe DMCA significantly reduces communication compared to broadcast.}

In Figure~\ref{fig:hist} we plot a 2D histogram showing the average communication link value for different X and Y positions of the neighboring agent on the ego frame. A communication link value of $1$ means a communication link is used, while $0$ means no communication link. We generate the 2D histogram by sampling positions and communication link predictions from circle scenario simulation with up to 10 agents. We can see DMCA-LC chooses the communicate with neighbors having ego frame position closer to origin. This also results in lower number of communication links in the case of DMCA-LC as seen in Table~\ref{tab:comm_link}.

{\color{black}An interesting behavior we observe with DMCA, DMCA-LC is that the robot avoids navigating through their neighbors' goal. This is illustrated through the Figure~\ref{fig:goal-aversion}. Through our end-to-end learning of communication and navigation, it appears the network has learnt to uses the communicated goal information from neighbors' to plan around the goal location. It is worth noting our reward structure does not explicitly reward this behavior.}

\section{Conclusion, Limitation, and Future Work}

{\color{black}We proposed a reinforcement learning approach for decentralized multi-robot navigation in complex scenarios. Our method learns to selectively communicate with neighboring agents and exchange their hidden state information to enhance navigation performance. We use a self-attention model to encode neighbor observations and perform link prediction for inter-agent communication. Our approach was shown to outperform other learning-based baselines in simulations across multiple scenarios. As a future work, we aim to analyze the trade-offs between communication efficiency and navigation performance, specifically considering the negative rewards for communication. Additionally, we plan to explore `what to communicate' to further improve navigation.}



\bibliographystyle{IEEEtran.bst}
\bibliography{IEEEabrv,references}

\end{document}